\documentclass[]{IEEEtran}
\ifCLASSOPTIONcompsoc
\else
  \usepackage{cite}
\fi

\usepackage[cmex10]{amsmath}
\usepackage{algorithmic}
\usepackage[boxed,lined]{algorithm2e}
\usepackage{graphicx}
\usepackage{color}
\usepackage{array}

\ifCLASSOPTIONcompsoc
\usepackage[tight,normalsize,sf,SF]{subfigure}
\else
\usepackage[tight,footnotesize]{subfigure}
\fi

\usepackage{url}

\newcommand{\setIV}{\mathcal V}

\newcommand{\ie}{i.~e.}
\newcommand{\Ie}{I.~e.}
\newcommand{\eg}{e.~g.}

\renewcommand{\vector}[1]{{\mbox{\boldmath$#1$}}}
\newcommand{\q}[1]{{{\bf #1}}}
\newcommand{\m}[1]{{\mbox{{\fontencoding{T1}\sffamily\slshape{#1\/}}}}}

\newcommand{\mR}{{\rm I\!R}}
\newcommand{\trans}[0]{^{\sf T}}

\newcommand{\new}[0]{_{\Delta}}
\newcommand{\inv}[0]{^{-1}}
\newcommand{\itero}[0]{_{\left( i-1\right)}}
\newcommand{\iter}[0]{_{\left( i\right)}}
\newcommand{\stitero}[0]{_{\left( s-1\right)}}
\newcommand{\stiter}[0]{_{\left( s\right)}}

\newcommand{\dimf}{{M}}

\newcommand{\class}{{\cal C}}  % Klasse

\newcommand{\vfeat}{\vector x} % Merkmalsvektor
\newcommand{\hfeat}{\q x} % erweiteter Merkmalsvektor
\newcommand{\mfeat}{\m X} % Merkmalsvektor

\newcommand{\vkernel}{\vector k} % kernel vector
\newcommand{\kernel}{\m K} % kernel matrix

\newcommand{\z}{\vector z}

\newcommand{\vparam}{\vector w}

\newcommand{\hparam}{\q w}

\newcommand{\valpha}{\mbox{\boldmath$\alpha$}}

\newcommand{\prob}{p}
\newcommand{\vprob}{\vector p}
\newcommand{\mprob}{\m P}

\newcommand{\mclass}{\m C}

\newcommand{\vlabels}{\vector y}
\newcommand{\target}{t}
\newcommand{\vtarget}{\vector t}

\newcommand{\uni}{\textsc{University of Pavia}\xspace}
\newcommand{\city}{\textsc{Center of Pavia}\xspace}
\newcommand{\indian}{\textsc{Indian Pines}\xspace}

\begin{document}

\title{Incremental Import Vector Machines for Classifying Hyperspectral Data}

\author{Ribana Roscher, 
        Bj\"orn Waske,~\IEEEmembership{Member,~IEEE,}
        Wolfgang F\"orstner,~\IEEEmembership{Member,~IEEE}% <-this % stops a space
\IEEEcompsocitemizethanks{\IEEEcompsocthanksitem The authors are with the Institute of Geodesy and Geoinformation,
Faculty of Agriculture, University of Bonn, 53115 Bonn, Germany (e-mail: rroscher@uni-bonn.de, bwaske@uni-bonn.de, wfoerstn@uni-bonn.de).\protect\\
This work was supported by CROP.SENSe.net project, funded by the German Federal Ministry of Education and Research (BMBF) within the scope of the competitive grants program Networks of excellence in agricultural and nutrition research (FKZ: 0315529).
}% <-this % stops a space
\thanks{Manuscript received May 4, 2011}}

\markboth{IEEE Transactions on Geoscience and Remote Sensing,~Vol.~50, No.~09, ~September 2012}%
{Roscher \MakeLowercase{\textit{et al.}}: Import Vector Machines for Classifying Hyperspectral Data}

\IEEEcompsoctitleabstractindextext{%
\begin{abstract}
%\boldmath
In this paper we propose an incremental learning strategy for import vector machines (IVM), which is a sparse kernel logistic regression approach.
We use the procedure for the concept of self-training for sequential classification of hyperspectral data.
The strategy comprises the inclusion of new training samples to increase the classification accuracy and the deletion of non-informative samples to be memory- and runtime-efficient. Moreover, we update the parameters in the incremental IVM model without re-training from scratch.
Therefore, the incremental classifier is able to deal with large data sets. 
The performance of the IVM in comparison to support vector machines (SVM) is evaluated in terms of accuracy and experiments are conducted to assess the potential of the probabilistic outputs of the IVM.

Experimental results demonstrate that the IVM and SVM perform similar in terms of classification accuracy. 
However, the number of import vectors is significantly lower when compared to the number of support vectors and thus, the computation time during classification can be decreased.
Moreover, the probabilities provided by IVM are more reliable, when compared to the probabilistic information, derived from an SVM's output. 
In addition, the proposed self-training strategy can increase the classification accuracy.
Overall, the IVM and the its incremental version is worthwhile for the classification of hyperspectral data.
\end{abstract}

}

\maketitle
\IEEEdisplaynotcompsoctitleabstractindextext
\IEEEpeerreviewmaketitle

%%%%%%%%%%%%%%%%%%%%%%%%%%%%%%%%%%%%%%%%%%%%%%%%%%%%%%%%%%%%%%%%%%%%%%%%%
% -------------------------------------------------------%
\section{Introduction}
Hyperspectral imaging, also known as imaging spectroscopy is used for more than two decades for monitoring the earth~\cite{Goetz2009}. 
The spectrally continuous data ranges from visible to the short-wave infrared region of the electromagnetic spectrum and thus, enables a detailed separation of similar surface materials.
Therefore hyperspectral imagery is used for classification problems that require a precise differentiation in spectral feature space~\cite{Waske2010, Mitri2010, Benediktsson2005}. 
Hyperspectral applications become even more attractive, regarding the increased availability of hyperspectral imagery through future space-borne missions, such as the German EnMAP (Environmental Mapping and Analysis Program)~\cite{Guanter2009} and the Italian PRISMA (Hyperspectral Precursor of the Application Mission).

Nevertheless, the special properties of hyperspectral imagery demand more sophisticated image (pre-)processing and analysis~\cite{Richards2005,Plaza2009}.
Conventional methods, such as the maximum likelihood classifier, can be limited when applied to hyperspectral imagery, due to the high-dimensional feature space and a finite number of training samples.
Consequently, the classification accuracy often decreases with an increasing number of bands (\ie, the well-known Hughes phenomena).
Thus, more flexible classifiers, such as spectral angle mapper, neural networks and support vector machines (SVM), are applied on hyperspectral imagery~\cite{Waske2010, Chen2007, VanderLinden2007, Benediktsson2005}.

Among the various developments in the field of pattern recognition, SVM~\cite{Vapnik2000} are perhaps the most popular approach in recent hyperspectral applications~\cite{Plaza2009}. 
SVM can outperform other methods in terms of the classification accuracy~\cite{Pal2006, Waske2009} and still exhibit further modification and improvement, \eg, in context of modifying the kernel functions~\cite{Camps2010} and semi-supervised learning~\cite{Mari2010}.
Whereas other classifiers can directly solve multi-class problems, the binary nature of SVM requires an adequate multi-class strategy~(\eg,~\cite{Mathur2008}).
In contrast to other classifiers, which directly provide class labels or probabilities, SVM provide the distance of each pixel to the hyperplane of the binary classification problem. This information is used to determine the final class membership.
Although the output of SVM can be transferred to probabilities~(\eg, \cite{Platt2000}), the reliability of these values could be inadequate \cite{Foody2008, Tipping2001}. Nevertheless, probabilities are of interest and can be used \eg, as input in a markov random field model or for uncertainty analysis \cite{Giacco2010, Zhong2010, Tarabalka2010}.

Logistic regression as an alternative probabilistic discriminative classification model was already used in context of classification and feature selection of hyperspectral imagery \cite{Zhong2008,Cheng2006}.
The approach was extended to kernel logistic regression, \eg, \cite{Keerthi2005, Cawley2004}, showing a better accuracy but a higher complexity.
To overcome the limitations in context of efficiency and computation time several sparse realizations of (kernel) logistic regression have been developed, including the explicit usage of a sparsity enforcing prior \cite{Borges2011,Li2010,Cawley2007,Krishnapuram2005}, an implicit prior used in the relevance vector machines (RVM) \cite{Foody2008,Demir2007,Tipping2001} or a greedy subset selection for the concept of import vector machines (IVM) \cite{Zhu2005}.

Recent classifier developments, such as SVM, usually perform well on high-dimensional data sets. 
Nevertheless, the classification accuracy can be affected when a limited number of training samples is available~\cite{Waske2010,Pal2010}.
One possibility to overcome this problem is active learning, which has been studied extensively in the literature, \eg, \cite{Tuia2011,Li2010,Li2010a,Rajan2008}.
In this paper we use the specialized concept of self-training \cite{Ng2003}.
Self-training is based on the sequentially training of a classifier with new training samples.
\Ie, starting with a initial classifier model learned from a few labeled training samples, an iterative classification procedure is performed.  
After each classification, new relevant training samples are selected and usually the classifier is re-trained from scratch.
That means, the classifier is learned by using old and new training samples without taking into account the previous learned classifier model.
For the acquisition of new training samples we use spectral as well as spatial information by using a discriminative random field (DRF) \cite{Kumar2006}.
The gain of integrating spatial information was already discussed in several studies in context of classifying hyperspectral data~\cite{Borges2011,Tarabalka2010,Li2010a,Zhong2010,Zhong2008a}. 
Moreover, we consider the probabilistic outputs by the IVM, to assess the reliability of the classification result.

However, to be time- and memory-efficient we need an incremental learning strategy within the self-training approach.
Several incremental learning methods have been proposed, extending classical and
state-of-the art off-line methods.
Incremental generative models, \eg \cite{Fei-Fei2007}, provide
probabilities but tend to become very complex.
Discriminative models, on the other hand, like incremental linear discriminant
analysis \cite{Pang2005} or incremental support vector machines, \eg
\cite{Cauwenberghs2001}, show a good performance in classification tasks.

Kernel-based algorithms have achieved considerable success in incremental and
off-line learning settings \cite{Karasuyama2010, Fung2002}.
However, incremental kernel-based learning settings need strategies for dealing
with existing and new samples.
The challenge is to perform efficient, incremental update steps without
suffering a loss in performance.

The main objective is to propose an incremental learning strategy to update the trained IVM model.
The classifier is called incremental IVM.
Therefore, the IVM model is not re-trained from scratch, as usually done in context of active and self-training approaches. 
The incremental IVM consists of the selection of new training samples, the deletion of irrelevant training samples, and the update of the IVM model. 
Therefore, the classifier is able to deal with infinitely new training samples. 

The potential of the IVM and the its incremental version in context of classifying hyperspectral data is evaluated.
The performance is compared to SVM in terms of accuracy.
In addition, the reliability of the probabilities outputs provided by IVM and SVM are evaluated by using a discriminative random field and the effect is further investigate by neglecting uncertain test samples.
Our study is aiming on the classification of three different hyperspectral data sets, \ie, two urban areas from the city of Pavia and an agricultural area from Indiana, USA, using SVM and IVM.

The paper is organized as follows. 
Section~\ref{sec:background} discusses the logistic regression, kernel logistic regression and the IVM algorithm. 
Moreover, the concept of SVM and related classifiers is briefly introduced and compared to IVM. 
Section~\ref{sec:iivm} introduces the proposed strategy for self-training, including the DRF model.
Also the incremental IVM are explained.
The experimental setup is given in Section~\ref{sec:experiments}. 
The results are presented and discussed in Section~\ref{sec:discussion}. 
We conclude in Section~\ref{sec:conclusion}.

% -------------------------------------------------------%
\section{Theoretical Background}
\label{sec:background}
In this section IVM model is introduced.
Starting from the logistic regression, we discuss kernels and sparsity and finally the sparse kernel logistic regression model, i.e., IVM.
Also a brief introduction to SVM and sparse multinomial logistic regression is given.
In \ref{sec:drf} the discriminative random field is introduced, which incorporates spatial information.

\subsection{Logistic Regression and Kernel Logistic Regression}
\label{sec:logreg}
\paragraph{Logistic Regression}
We assume to have a training set $(\vfeat_n, y_n)$, $n = 1,\dots,N$ of $N$
labeled samples with feature vectors $\vfeat_n\in\mR^{\dimf}$ and class
labels $y_n \in \class = \{\class_1,\dots, \class_K\}$.
The observations are collected in a matrix
$\mfeat = \left[\vfeat_1, \ldots, \vfeat_N\right]$, while the corresponding
labels are summarized in the vector $\vlabels = \left[y_1, \ldots, y_N\right]$.

In the two-class case the posterior probability $\prob_n$ of a feature vector
$\vfeat_n$ is assumed to follow the Logistic Regression model
\begin{equation}
	\prob_n = \prob(y_n = \class_1|\hfeat_n; \hparam) = \frac{1}{1 + \exp(-\hparam\trans \hfeat_n)}
  \label{eq:probs}
\end{equation}
with the extended feature vector $\hfeat\trans_n = [1,\vfeat_n\trans] \in
\mR^{\dimf+1}$ and the extended parameters $\hparam\trans =
[w_{0},\vparam\trans] \in \mR^{\dimf+1}$ containing the bias $w_{0}$ and the
weight vector $\vparam$.

\paragraph{Kernel Logistic Regression}
\label{sec:KLR}
In linear non-separable cases, the original observations $\mfeat$ are implicitly mapped
from the input space to a higher-dimensional kernel space with the kernel matrix $\kernel = [k_{nm}]$
via the kernel function $k_{nm} = k\left(\hfeat_n, \hfeat_m\right)$.
The kernel matrix $\kernel$ consists of affinities between the points depending on the distance measure defined by the kernel function.

%% parameters and objective function
The parameters, in the kernel-based approach referred to $\valpha$, are determined in an iterative way with
\begin{align}
	\valpha\iter &=
    \left(\frac 1N \kernel \trans \m R \kernel
    + \lambda \kernel\right)\inv \kernel\trans \m R \vector z
    \label{eq:valpha1} \\
	\vector z &= \frac 1N \left(\kernel \valpha\itero
    + \m R\inv \left(\vprob - \vtarget\right)\right)
    \label{eq:ztilde1}
\end{align}
by optimizing the objective function
\begin{align}
	\mathcal{Q}\iter & = -\frac1N \sum_n\left[
      \target_{n} \log \prob_n + \left(1 - \target_n\right) \log \left(1 - \prob_n\right)\right]\nonumber\\
      &\hspace{1.2em} +\frac{\lambda}{2} \valpha\iter\trans \kernel \valpha\iter
	\label{eq:unreg}
\end{align}
using the Newton-Raphson procedure.
The $\left(N\times N \right)$-dimensional diagonal matrix $\m R$ has the
elements $r_{nn} = p_n\left(1 - p_n\right)$ with $\prob_n = 1 / (1 + \exp(-\vkernel_{n} \valpha))$ and $\vkernel_{n}$ as the $n$th row of the kernel matrix $\kernel$.
The binary target vector $\vtarget\in\{0,1\}$ of length $N$ codes the labels
with $\target_n = 0$ for $y_n = \mathcal C_1$ and $\target_n = 1$ for $y_n = \mathcal C_2$.
Additionally, we add an $L_2$-norm regularization term with parameter $\lambda$
to prevent overfitting.

\subsection{Import Vector Machines}
\label{sec:ivm}
The kernel logistic regression includes all training samples to train the classifier,
which is computationally expensive and memory intensive for data sets with many training samples.
Similar to the SVM the IVM algorithm \cite{Zhu2005} chooses a subset $\setIV$ of feature vectors out of the training set with $V = |\mathcal V|$ samples $\m X_{\setIV}=[\hfeat_{\mathcal V,m}]$, $m=1, \ldots, V$, obtaining a sparse solution of the kernel logistic regression.
These feature vectors are called import vectors.

Following (\ref{eq:valpha1}) and (\ref{eq:ztilde1}) the parameters in iteration $i$ are determined by
\begin{align}
	\valpha\iter &=
    \left(\frac 1N \kernel_{\mathcal V} \trans \m R \kernel_{\mathcal V} 
    + \lambda \kernel_R\right)\inv \kernel_{\mathcal V} \trans \m R \vector z
    \label{eq:valpha} \\
	\vector z &= \frac 1N \left(\kernel_{\mathcal V}  \valpha\itero
    + \m R\inv \left(\vprob - \vtarget\right)\right).
    \label{eq:ztilde}
\end{align}
The $(N\times V)$-dimensional kernel matrix is given by $\kernel_{\mathcal V} = [k(\hfeat_n,
\hfeat_{\mathcal V,m})]$ and the $(V\times V)$-dimensional regularization matrix by
$\kernel_R = [k(\hfeat_{\mathcal V,l}, \hfeat_{\mathcal V,m})]$, $\{l,m\}=1, \ldots, V$.

The IVM is illustrated in Algorithm~\ref{alg:ivm}.
\begin{algorithm}[ht]
  Initialize
    $\mathcal V_0 := \{\}$,
    $\mathcal X_0 := \{\vfeat_1, \ldots, \vfeat_N\}$,
    $i := 0$\;
  \Repeat{$\mathcal Q$ \rm{converged}}{
    Compute $\vector z\iter$ from the current set $\mathcal V\iter$\;
    \ForEach {$\vfeat_n \in \mathcal X\iter$}{
      Let $\mathcal V_{(i)n} := \mathcal V\iter \cup \vfeat_n$\;
      Compute $\valpha_{(i)n}$ from $\mathcal V_{(i)n}$ in a one-step iteration\;
      Evaluate error function $\mathcal Q_{(i)n}$\;
    }
    Find best point $\vfeat^* = \vfeat_n$ with
      $n = \operatorname{argmin}_n \mathcal Q_{(i)n}$\;
    Update
      $\mathcal V_{(i+1)} := \mathcal V\iter \cup \vfeat^*$,
      $\mathcal X_{(i+1)} := \mathcal X\iter \setminus \vfeat^*$,
      $i := i+1$\;
    %Recompute $\valpha_i$, $\vprob_i$ and $\mathcal Q_i$\;
  }
  \caption{
    IVM:
    In every iteration $i$ each point
    $\vfeat_n \in \mathcal X\iter$ from the current training set
    $\mathcal X\iter$
    is tested to be in the set of import vectors $\mathcal V\iter$.
    The point $\vfeat^*$ yielding the lowest error $\mathcal Q_{(i)n}$ is
    included.
    The algorithm stops as soon as $\mathcal Q$ converged.
  }
  \label{alg:ivm}
\end{algorithm}
The convergence criterion is proposed by the ratio
$\epsilon = |\mathcal Q\iter - \mathcal Q_{\left(i - \Delta i\right) }| / |\mathcal Q\iter|$ with
a small integer $\Delta i$.

The original algorithm selects the import vectors in a greedy forward selection procedure. 
The approach is extended to a forward stepwise selection, which allows forward and backward steps.
The advantage of this procedure is, that import vectors, which once entered can be dropped if they are no longer relevant. 
In all experiments an improvement of the results could be observed.
Furthermore, an incremental update procedure to compute the inverse in (\ref{eq:valpha}) depending on the last iteration is used, which makes the algorithm more efficient.
The incremental update is described in a detailed way in Section \ref{sec:incrementalLearning}.

%-------------------------------------------------------------------------
The two class model can be generalized to the multi-class model.
Then the objective function is
\begin{equation}
	\mathcal Q = -\frac{1}{N} \sum\limits_n \vtarget\trans_n \log \vprob_n
    +  \frac{\lambda}{2} \sum\limits_k \valpha_k\trans \kernel_R \valpha_k
\end{equation}
with the probabilities $\mprob = \left[ \prob_1, \dots, \prob_N \right]$ obtained by
\begin{equation}
	\prob_{nk} = \frac{\exp(\vkernel_{\mathcal V,n} \valpha_k)}
    {\sum_l \exp(\vkernel_{\mathcal V,n} \valpha_l)}.
\end{equation}
The binary target vector $\vtarget_n$ of length $K$ uses the 1-of-K coding scheme
so that all components but $t_{nk}$ are $0$ if the point
$\vfeat_n$ is from class $\mathcal C_k$. 
In the Newton-Raphson procedure in (\ref{eq:valpha}) and (\ref{eq:ztilde}) we have to use 
one $\m R_k$ and one $\vector z_k$ for each class.

%-------------------------------------------------------------------------
\subsection{Related Classifiers}
Recently several algorithms have been developed, which enforce sparseness to control both the generalization capability of the learned classifier model and the complexity, as \eg, \cite{Figueiredo2003, Figueiredo2001, Csato2002, Lawrence2003}.
In these algorithms the model consists of a sparse weighted linear combination of basis functions, which are the input features themselves, nonlinear transformations of them or kernels centered on them.
In this section we restrict ourselves to the review of realizations of sparse (kernel) logistic regression and SVM, whereby the latter one is used for comparison in the experiments.

\subsubsection{Realizations of Sparse (Kernel) Logistic Regression}
\label{sec:sparseKLR}
Using logistic regression or its kernel realization can be prohibitive regarding memory and time requirements if the dimension of the features or the number of training samples is large.
Several sparse algorithms have been developed in the last years to overcome this problem.

The relevance vector machine \cite{Tipping2001} uses the same model as the kernel logistic regression in combination with an implicit prior as regularization term, the so-called ARD (automatic relevance determination) \cite{Neal1996} prior, to induce sparseness.
The prior includes several regularization parameters, also called hyperparameters, which are determined during the optimization process.
The algorithm have shown to be very sparse, but also tends to underfit, leading to a non well-generalized model \cite{Krishnapuram2005}.
Additionally, the RVM uses an expectation-maximization (EM)-like learning method and therefore, can suffer from local minima leading to non-optimal classification results.

Alternatively, \cite{Cawley2007, Krishnapuram2005} use a Laplace prior enforcing sparseness, which assigned regularization parameter is determined via cross-validation.
These approaches propose sparse multinomial (kernel) logistic regression (SMLR) using different methods for a fast computation.
In the field of hyperspectral image classification these approaches have been applied and further developed in, \eg, \cite{Borges2011,Li2010}.

\subsubsection{Support Vector Machines}
\label{sec:SVM}
The SVM find an optimal nonlinear decision boundary by minimizing the objective function
\begin{equation}
	\mathcal Q_{SVM} = \frac{1}{N}\sum\limits_n \left[1-y_n f(\vector x_n) \right]_+ + \frac{\lambda}{2}\Vert f \Vert^2
\end{equation}
with $f(\vector x_n) = \sum_n \alpha_n\kernel(\m X, \vector x_n)$. 

Contrary to IVM, which maximize the posterior probabilities, SVM aim to maximize the margin between the hyperplane and the closest training samples, the so-called support vectors. SVM are a binary classifier, with the decision rule given by the sign of $f(\vector x_n)$.

\subsubsection{Comparison with Import Vector Machines}
The main properties of SVM, IVM, SMLR and RVM are summarized in Table \ref{tab:summary} and briefly compared below. 

\begin{table*}[ht]
\caption{Summary of the SVM, RVM and IVM algorithm regarding several characteristics. Here "--/0/+" means that the algorithm satisfies a certain property 
"barely/partially/completely".}
\centering
\begin{tabular}{l|l|c|c|c|c|c|}
	Algorithm & Objective function, optimization procedure & Sparse & Training & Testing & Probabilistic & Multi-class\\
	&  &  & time & time &  &\\
	\hline	
	SVM & convex, greedy with SMO algorihtm    & 0 & +  & 0 & 0 & 0\\
	RVM & nonconvex, IRLS, EM-like & + & --  & + & + & +\\
	SMLR & convex, see Sec. \ref{sec:sparseKLR} & 0/+ & 0/+ & 0/+ & + & + \\
	IVM & convex, IRLS with greedy forward stepwise & + & 0/+ & + & + & +\\
	& selection (see Sec. \ref{sec:ivm} and \ref{sec:incrementalLearning}) & &  & & &\\
\end{tabular}
\label{tab:summary}
\end{table*}

The objective function of the SVM is quadratic and therefore convex and can be efficiently solved with sequential minimal optimization algorithm (SMO) \cite{Platt1999}.
The objective function of the IVM is convex, but non-quadratic. 
The function can be solved with iterated re-weighted least squares (IRLS) and a greedy forward selection of import vectors.

All models are sparse, i.e., the model parameters are primarily zero. 
The sparseness arises from different techniques, namely the usage of a prior in the SMLR models and the RVM or the greedy selection of a fraction of the training samples in the IVM model.
However, IVM and RVM have shown to be sparser when compared to SVM \cite{Demir2007, Zhu2005}, and thus, require less computation time during classification. 
The training time of the IVM, on the other hand, can be slower than for the SVM, because of the non-quadratic objective function. 
Nonetheless, the training time depends on the number of training samples and the attainable sparseness, that therefore also SMLR and IVM can reach a faster training time.

In comparison to RVM and SMLR approaches, which use the whole kernel during the optimization procedure, the IVM algorithm is suitable for large data sets, since only a subset of the training samples is used for computations during the optimization. 
Consequently, also only a fraction of the whole kernel have to be computed and stored. 
In contrast to this, \eg, the standard RVM algorithm has to solve a large matrix inversion of the size of the number of training samples during the optimization process and therefore can be slow and intractable for large datasets. 

While IVM, SMLR and RVM directly provide a probabilistic output, the SVM algorithm has the opportunity to transform its output to probabilities after some post-processing steps~\cite{Platt2000}.
Though, as Tipping \cite{Tipping2001} already shows the transformed output is not necessarily statistical interpretable.

Both IVM, SMLR and RVM are introduced as multi-class classifiers. 
The standard SVM have the opportunity to multi-class classification but they need a coupling strategy to get multi-class classification results. 
Moreover, these multi-class approaches are more suitable for practical use than direct multi-class approaches for SVM \cite{Hsu2002}.
%the reliability of the values can be questioned [18],[20].

\subsection{Discriminative Random Fields}
\label{sec:drf}

A discriminative random field is employed to model prior knowledge about the neighborhood relations within the image.
The final classification is assumed to be smooth,
\ie, neighboring pixels are more likely to belong to the same class than to different classes. 

With this, the best classification $\mclass_{\text{DRF}}$ is given by the argument of the minimum of the energy
\begin{equation}
	E\left(\vector y\right) =-\sum\limits_{j \in \mathcal I}\log \prob\left( y_j |\vfeat_j\right) -
    \beta \sum\limits_{\{m,j\} \in \mathcal N} \delta \left( y_j,y_m\right),
	\label{eq:DRF}
\end{equation}
where $\vfeat_j$ is the observed feature vector from the $j$th pixel,
$\mathcal I$ being the set of
all pixels and $\delta$ being the Kronecker delta function.
The weighting parameter is given by $\beta$, which can be determined via cross-validation.
The first term in~(\ref{eq:DRF}) models the probability of a class assignment $y_j$ of the $j$th pixel, defined by the probabilistic output of the IVM.
The second term describes the interaction potential as a Potts model over a 2D
lattice penalizing every dissimilar pair of labels and therefore heterogeneous regions.
The set of all neighboring pixels is given by $\mathcal N$.

% -------------------------------------------------------%
\section{Incremental Learning Strategy for Self-Training}
\label{sec:iivm}
In this section we introduce the self-training concept and the learning strategy for the incremental IVM.
Self-training refers to the sequential selection of new training data and the adaption of the previous classifier model.
The previous classifier model can (i) be neglected and re-trained from scratch or (ii) incrementally be updated.
Following (i), a regular classifier training is performed, using the whole training samples set (\ie, previous + new training samples). The latter approach "simply" updates the previous model, using the newly selected training samples.

\subsection{Self-training Concept}
\label{sec:self}
\begin{figure}[ht]
  \centering
   \includegraphics[width=1\columnwidth]{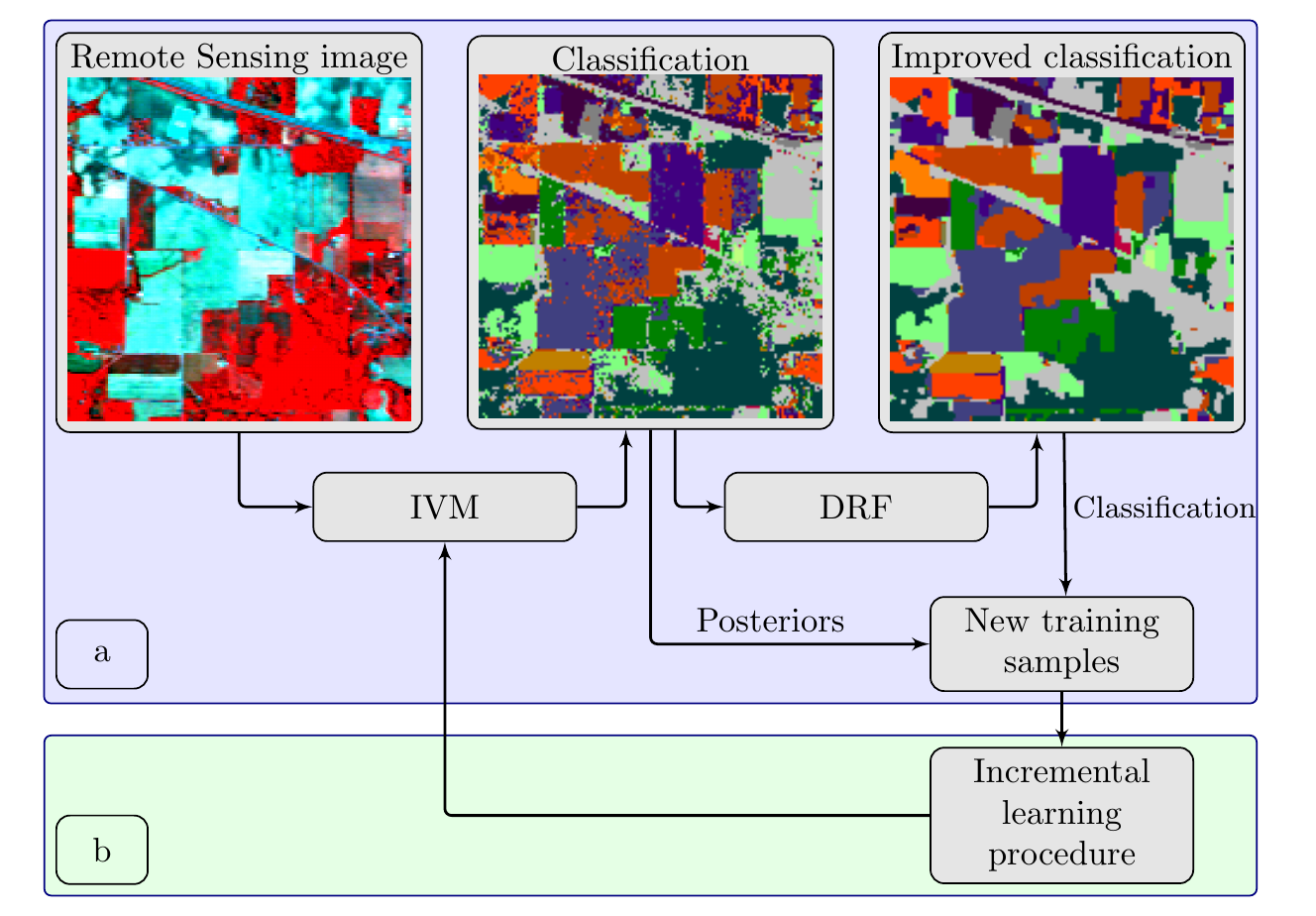}
	\caption{Self-training scheme consisting of two steps: In the first step (Fig. 1a) the image is classified with the learned incremental IVM model and the DRF. The probabilistic output of the IVM and the classification result of the DRF are used to acquire new training samples. In the second step (Fig. 1b) the classifier model is incrementally updated.}
	\label{fig:self}
\end{figure}

In the first step (Fig. 1a) we identify potential new training samples.
For this selection step, we use the classification result provided by the DRF and the probabilistic outputs of the incremental IVM. 
In the second step (Fig. 1b) we use the incremental learning strategy to update the classifier model, without re-training from scratch. The procedure is repeated until no additional training samples can be selected.
The latter step consists of the inclusion of new training samples, the deletion of irrelevant training samples and finally, the update of the IVM model. Therefore the approach can handle large data sets or even infinitive data streams. However, this step is independent of the used self-training strategy and other approaches can be used to identify new training samples.
Both steps are explained in detail in the next paragraphs.

\subsection{Acquisition of New Training Samples for Self-training}
\label{sec:acquisition}
\begin{figure*}[ht]
  \centering
   \includegraphics[width=0.6\textwidth]{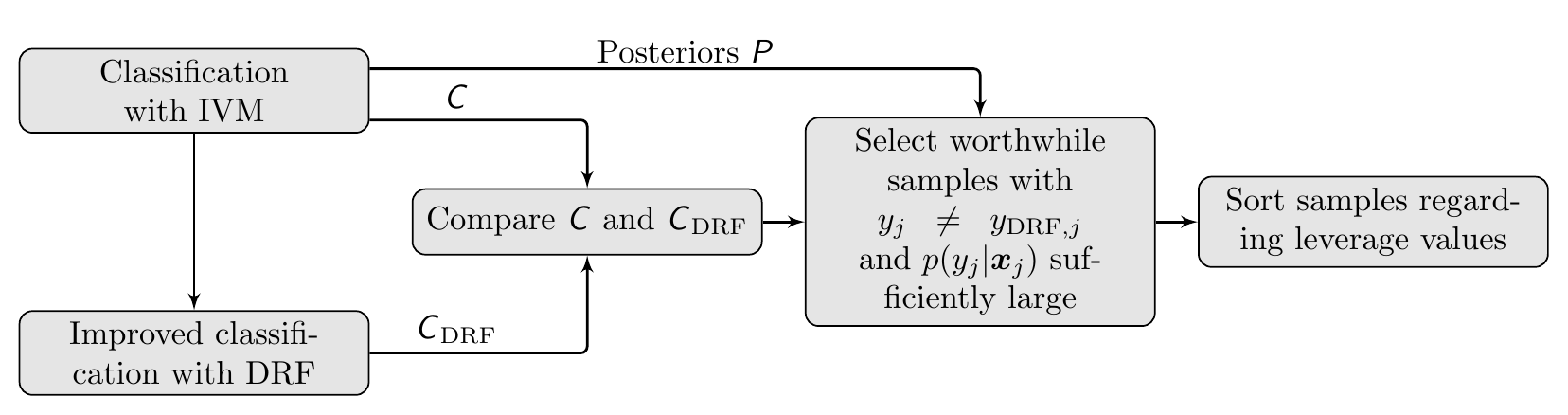}
	\caption{Schematic diagram of the selection of new training samples: Worthwhile samples are identified by comparing the classification $\mclass$ and $\mclass_{\text{DRF}}$. Samples with a relatively small posteriori probability are excluded. Remaining samples are sorted regarding their leverage value, \ie, samples with high influence are preferred.}
	\label{fig:acquisition}
\end{figure*}

This acquisition step illustrated in Fig. \ref{fig:acquisition} is subdivided in 3 parts:
First, we use a DRF as expert to identify worthwhile samples by evaluating the disagreement between the classification $\mclass$ yielded by the IVM classifier and the DRF result $\mclass_{\text{DRF}}$. The aspect implies, that the samples with estimated label $y_{\text{DRF},j}$ derived from the DRF are sufficient uncertain with $\max(\vprob_j) < 0.5$ obtained from the IVM classifier. Using such samples we ensure progress in self-training. 

Second, we exclude from the selected samples these ones, whose probability is too small. \Ie, the influence of the new samples is restricted, to ensure that the model is gradually changed and stable.

Finally, we sort the chosen samples by their potentially influence on the model, to enable the flexibility of the model during self-training. The potential influence uses the concept of leverage points in regression~\cite{Rousseeuw1987}.
The leverage values in a weighted regression are contained in the vector
\begin{equation}
  \vector l = \text{diag}\left(
    \kernel_{\text{pot}}\left(\kernel_{\text{pot}}\trans \m R_{\text{pot}} \kernel_{\text{pot}}\right)\inv
    \kernel_{\text{pot}}\trans\m R_{\text{pot}}\right)\,.
\end{equation}
The kernel matrix obtained from the potential new training samples $\mfeat_{\text{pot}}$ is given by $\kernel_{\text{pot}} = [k(\hfeat_{\text{pot},j},
\hfeat_{\mathcal V,m})]$ and $\m R_{\text{pot}}$ is the weight matrix of the class the training sample belongs to.
Training samples with a high leverage value first are considered first, since they causes large effects in the learned model. 
The self-training is stopped if no more training samples can be acquired.

To prevent an acquisition, which leads to an imbalanced number of training samples per class, we ensure sampling an equal number of training samples for each class. If not enough new training samples can be acquired with the proposed self-training approach, we also consider training samples with a high probability and whose label in $\mclass$ and $\mclass_{\text{DRF}}$ are the same.
These samples have a small influence onto the model and should not change the result too much, but balance the number of samples of each class.
If there are still not enough samples for a class, we follow the over-sampling approach by duplicating existing samples from the concerned class at random and add a small noise to them \cite{Japkowicz2000}.

\subsection{Incremental Learning}
\label{sec:incrementalLearning}
To update the learned classifier, we consider the following aspects:
\begin{itemize}
	\item New training samples acquired with the proposed self-training approach (see Section \ref{sec:acquisition}) are included.
	\item Non-informative samples are deleted.
	\item The set of import vectors is updated.
\end{itemize}
 
\paragraph{Update Training Vectors} 
For the two-class case the incremental learning procedure is stated as follows.
We add training vectors $\mfeat\new$ with targets $\vtarget\new$ so that
$N\stiter := N\stitero + \Delta N$ with $\Delta N$ as the number of new training samples.
At each self-training iteration~$s$ we extend the matrices and vectors
\[\kernel\stiter =
    \left[\begin{array}{c}
      \kernel\stitero \\
      \kernel\new \\
    \end{array}\right],
\vtarget\stiter =
    \left[\begin{array}{c}
      \vtarget\stitero \\
      \vtarget\new \\
    \end{array}\right]
\]     
to obtain the updated parameters $\valpha\stiter^+$ given by (\ref{eq:inverse}) and (\ref{eq:z}) yielding
\[   
%\m R \stiter =
%    \left[\begin{array}{cc}
%      \m R\stitero & \\
%      & \m R\new \\
%    \end{array}\right],\,
\z\stiter =
    \left[\begin{array}{c}
      \z\stitero \\
      \z\new \\
    \end{array}\right],\,
 \vprob\stiter =
    \left[\begin{array}{c}
      \vprob\stitero \\
      \vprob\new \\
    \end{array}\right].
\] 
\begin{figure*}
\begin{align}
  \label{eq:inverse}
  \valpha\stiter^+ & =
  \left(1/N\stiter\left(\kernel\stitero \trans \m R\stitero \kernel\stitero +
    \kernel\new \trans \m R\new \kernel\new\right)+\lambda \kernel_{R,(s-1)}\right)\inv
    \left(\kernel\stitero\trans \m R\stitero \z\stitero +
    \kernel\new\trans \m R\new \z\new\right)\\
  \label{eq:z}  
  \z\stiter &= \frac1{N\stiter} \left[\begin{array}{c}
    \kernel\stitero \valpha\stitero +
    \m R\stitero\left(\vprob\stitero - \vtarget\stitero\right)\\
    \kernel\new\valpha\stitero  +
    \m R\new\left(\vprob\new-\vtarget\new\right) 
  \end{array}\right]
\end{align}
\end{figure*}

The Sherman-Morrison-Woodbury (SMW) \cite{Higham2002} formula is used to compute the inverse in
(\ref{eq:inverse}) yielding (\ref{eq:alphaSMW})
\begin{figure*}
\begin{equation}
  \valpha^+\stiter =
  A\inv - 1/N\stiter A\inv\kernel\new\trans
   \left(\m R\new\inv +1/N\stiter \kernel\new A\inv\kernel\new\trans\right)\inv
    \kernel\new A\inv 
  \left(\kernel\stitero\trans \m R\stitero \z\stitero +
    \kernel\new\trans \m R\new \z\new\right)
  \label{eq:alphaSMW}
\end{equation}
\end{figure*}
with
$A = 1/N\stiter \kernel\stitero \trans \m R\stitero \kernel\stitero + \lambda
\kernel_{R,(s-1)}$.
Note that the update (\ref{eq:alphaSMW}) only incorporates an inverse of size
$\Delta N \times \Delta N$, since the inverse $A\inv$ was computed in
time step $s-1$.
With these steps the parameters can be updated in an incremental way without re-training from scratch.

The update (\ref{eq:inverse}) can also be formulated for a decreasing number of
training vectors in a similar manner, leading to an efficient update rule as
in (\ref{eq:alphaSMW}), only with the sign of $\m R\new$ changed. 
Training vectors can be removed based on their "age", so that they follow the first-in-first-out strategy.
This can lead to instable results.
Therefore, we identify training vectors, which can be removed using Cook's distance~\cite{Cook1982}.
Cook's distance measures the effect of deleting a training vector as the higher the value the more informative is the training vector:
\begin{equation}
	d_n = \frac{\left(\vprob_n-\vtarget_n\right)\trans\left(\vprob_n-\vtarget_n\right)}{a~\text{MSE}} \frac{l_n}{(1-l_n)^2}
\end{equation}
with $a$ as the number of parameters and MSE as the mean squared error summed over all classes given by the difference between $1$ and the mean probability of all training samples belonging to class $\mathcal C_k$. 
The leverage value of each training vector is given by $l_n$.
The training vectors with the lowest distance are removed in a greedy backward selection until the value of the optimization function increases more than of 5\%.

\paragraph{Update the Set of Import Vectors}
To add and remove import vectors we use the forward stepwise selection as described in Section~\ref{sec:background}.
We also make use of the SMW formula and proceed in the same way as described in Algorithm~\ref{alg:ivm} until a
convergence criterion is reached.

To generalize the two class model to the multi-class model we use the class-specific values $\m R_{k,(s-1)}$ and $\vector z_{k,(s-1)}$.

% -------------------------------------------------------%
\section{Experimental Setup}
\label{sec:experiments}
\definecolor{asphalt}{rgb}{0.501960784313726,0,0}
\definecolor{bare}{rgb}{0.250980392156863,0,0}
\definecolor{bitumen}{rgb}{0.752941176470588,0.501960784313726,0.501960784313726}
\definecolor{meadows}{rgb}{0,0.7529411764705880,0}
\definecolor{bricks}{rgb}{0.501960784313726,0,0.501960784313726}
\definecolor{shadow}{rgb}{0,0.501960784313726,0.501960784313726}
\definecolor{tiles}{rgb}{0.501960784313726,0.501960784313726,0.501960784313726}
\definecolor{trees}{rgb}{0,0.501960784313726,0}
\definecolor{water}{rgb}{0,0,0.501960784313726}

\definecolor{metal}{rgb}{0.250980392156863,0.250980392156863,0.250980392156863}
\definecolor{gravel}{rgb}{0.752941176470588,0.752941176470588,0.752941176470588}

\definecolor{alalfa}{rgb}{0.7529411764705880,0,0.250980392156863}
\definecolor{corn1}{rgb}{0.752941176470588,0.250980392156863,0}
\definecolor{corn2}{rgb}{1,0.250980392156863,0}
\definecolor{corn}{rgb}{1,0.501960784313726,0}
\definecolor{grass}{rgb}{0.501960784313726,1,0.501960784313726}
\definecolor{pasture}{rgb}{0,0.501960784313726,0}
\definecolor{hay}{rgb}{0.752941176470588,0.752941176470588,0.752941176470588}
\definecolor{soy1}{rgb}{0.250980392156863,0,0.5019607843137260}
\definecolor{soy2}{rgb}{0.250980392156863,0.250980392156863,0.5019607843137260}
\definecolor{soy3}{rgb}{0.250980392156863,0,0.250980392156863}
\definecolor{wheat}{rgb}{0.752941176470588,0.501960784313726,0}

\definecolor{woods}{rgb}{0,0.250980392156863,0.250980392156863}
\definecolor{bldg}{rgb}{0.752941176470588,0.752941176470588,0.752941176470588}
\definecolor{stone}{rgb}{0.501960784313726,0.501960784313726,0.501960784313726}
\definecolor{grass2}{rgb}{0.752941176470588,1,0.501960784313726}
\definecolor{oats}{rgb}{0.501960784313726,0.501960784313726,0.250980392156863}

\subsection{Data Sets}
We use three hyperspectral data sets -- \city, \uni and \indian -- from study sites with different environmental setting.
The data sets have been used in a multitude of studies, \eg, \cite{Yang2010, Camps2010, Pal2010, Plaza2009, Melgani2004}. 

The \city image was acquired by ROSIS-3 sensor in 2003.
The spatial resolution of the image is 1.3~meter per pixel.
The data cover the range from 0.43~$\mu$m to 0.86~$\mu$m of the electromagnetic spectrum.
However, some bands have been removed due to noise and finally 102~channels have been used in the classification.
The image strip, with 1096$\times$492 pixels in size, lies around the center of Pavia.
The classification is aiming on 9 land cover classes.
The \uni data set was also acquired by ROSIS-3 sensor with 610$\times$340 pixels in size and 103 channels.
The classification is aiming on 9 land cover classes.
The \indian data set was acquired by the AVIRIS instrument in 1992.
The study site lies in a predominately agricultural region in NW Indiana, USA.
AVIRIS operates from the visible to the short-wave infrared region of the electromagnetic spectrum, ranging from 0.4~$\mu$m to 2.4~$\mu$m.
The data set covers 145$\times$145 pixels, with a spatial resolution of 20~m per pixel.
The experiments are aiming on the classification of 16~classes (Table~\ref{tab:points}).

\begin{table*}[htb]
\centering
\caption{Number of training and test samples.}
\begin{tabular}{l|c|c|l|c|c|l|c|c}
\multicolumn{3}{c|}{\city} & \multicolumn{3}{c|}{\uni} & \multicolumn{3}{c}{\indian} \\
      class & \# training & \# test & class & \# training & \# test  & class & \# training & \# test  \\
\hline\hline
\colorbox{asphalt}{\parbox[h][0.2em][c]{1em}{}} Asphalt 			& 678 & 6907  & 
\colorbox{asphalt}{\parbox[h][0.2em][c]{1em}{}} Asphalt 			& 548 & 6304 & 
\colorbox{alalfa}{\parbox[h][0.2em][c]{1em}{}} Alalfa      		& 27  & 27\\
\colorbox{bare}{\parbox[h][0.2em][c]{1em}{}} Bare Soil 		& 820 & 5729  & 
\colorbox{bare}{\parbox[h][0.2em][c]{1em}{}} Bare Soil 		& 532 & 4572 & 
\colorbox{corn1}{\parbox[h][0.2em][c]{1em}{}} Corn-no till 	& 778 & 656\\
\colorbox{bitumen}{\parbox[h][0.2em][c]{1em}{}} Bitumen 			& 808 & 6479  & 
\colorbox{bitumen}{\parbox[h][0.2em][c]{1em}{}} Bitumen 			& 375 & 981  &
\colorbox{corn2}{\parbox[h][0.2em][c]{1em}{}} Corn-min till 	& 408 & 426\\
\colorbox{meadows}{\parbox[h][0.2em][c]{1em}{}} Meadows 			& 797 & 2108  & 
\colorbox{gravel}{\parbox[h][0.2em][c]{1em}{}} Gravel 			& 392 & 1815 & 
\colorbox{corn}{\parbox[h][0.2em][c]{1em}{}} Corn 			& 110 & 124\\
\colorbox{bricks}{\parbox[h][0.2em][c]{1em}{}} Bricks 			& 485 & 1667  & 
\colorbox{meadows}{\parbox[h][0.2em][c]{1em}{}} Meadows 			& 540 & 18146& 
\colorbox{grass}{\parbox[h][0.2em][c]{1em}{}} Grass/Pasture	& 208 & 289\\
\colorbox{shadow}{\parbox[h][0.2em][c]{1em}{}} Shadow 			& 195 & 1970  & 
\colorbox{metal}{\parbox[h][0.2em][c]{1em}{}} Metal Sheets 	& 265 & 1113 & 
\colorbox{pasture}{\parbox[h][0.2em][c]{1em}{}} Pasture/Trees 	& 339 & 408\\
\colorbox{tiles}{\parbox[h][0.2em][c]{1em}{}} Tiles 			& 223 & 2899  & 
\colorbox{bricks}{\parbox[h][0.2em][c]{1em}{}} Bricks 			& 514 & 3364 & 
\colorbox{hay}{\parbox[h][0.2em][c]{1em}{}} Hay 				& 228 & 261\\
\colorbox{trees}{\parbox[h][0.2em][c]{1em}{}} Trees 			& 785 & 5723  & 
\colorbox{shadow}{\parbox[h][0.2em][c]{1em}{}} Shadow		 	& 231 & 795  & 
\colorbox{soy1}{\parbox[h][0.2em][c]{1em}{}} Soybeans-no till	& 506 & 462\\
\colorbox{water}{\parbox[h][0.2em][c]{1em}{}} Water 			& 745 & 64533 & 
\colorbox{trees}{\parbox[h][0.2em][c]{1em}{}} Trees 			& 524 & 2912 & 
\colorbox{soy2}{\parbox[h][0.2em][c]{1em}{}} Soybeans-mid till	& 975 & 1493\\
- & - & -& - & - & - & 
\colorbox{soy3}{\parbox[h][0.2em][c]{1em}{}} Soybeans-clean till	& 265 & 349\\
 - & - & -& - & - & - & 
 \colorbox{wheat}{\parbox[h][0.2em][c]{1em}{}} Wheat 				& 100 & 112\\
- & - & -& - & - & - & 
\colorbox{woods}{\parbox[h][0.2em][c]{1em}{}}  Woods 				& 637 & 657\\
 - & - & -& - & - & - & 
 \colorbox{bldg}{\parbox[h][0.2em][c]{1em}{}} Bldg-Grass 		& 163 & 217\\
- & - & -& - & - & - & 
\colorbox{stone}{\parbox[h][0.2em][c]{1em}{}} Stone 					& 45 & 50\\
 - & - & -& - & - & - & 
 \colorbox{grass2}{\parbox[h][0.2em][c]{1em}{}} Grass/Pasture-mowed 	& 12 & 14\\
- & - & -& - & - & - & 
\colorbox{oats}{\parbox[h][0.2em][c]{1em}{}} Oats 					& 10 & 10\\
\hline
Total & 5536 & 91108 & & 3921 & 33698 & & 4811 & 5555
\end{tabular}
\label{tab:points}
\end{table*}

\subsection{Methods}
In the experiments the IVM for classifying hyperspectral data is analyzed.
In addition SVM were applied on the data sets.
SVM are perhaps the most popular approach in more recent applications and seems particularly advantageous when classifying high-dimensional data sets.
Thus, the method is regarded as a kind of benchmark classifier for comparison with new approaches.

Moreover, a DRF is applied on the respective probabilistic output. 
Besides as input for the DRF, we use the probabilistic outputs to analyze the uncertainty of the classification result.
We assess the reliability of the probabilities by rejecting uncertain test samples and deriving the classification accuracy on the non-rejected test points \cite{Giacco2010}.
The rejection rate is given by a threshold on the posterior probability, whereby the accuracy provided by SVM and IVM is reported as a function of the rejection rate in discrete intervals.

In addition, the incremental IVM is evaluated by applying the self-training approach on the three data sets in terms of the classification accuracy and sparsity. 

To investigate the impact of the number of training samples on the performance of the model (e.g., in terms of sparsity and accuracy) we use two different training sets, containing (i) all initial training samples and (ii) 10\% of each class (with a minimum of at least 10 samples per class).
For (ii), we performed a stratified random sampling, selecting 10\% of the samples of each class from the initial training set.
The final results were averaged.

For the SVM and the (incremental) IVM we use a radial basis function kernel.
The kernel parameters are determined by a 5-fold cross-validation.
Also the DRF parameter $\beta$ is determined by by 5-fold cross-validation.
The result provided by the common IVM is used for the initialization of the self-training.
The self-training procedure is repeated until no more training samples can be selected.

The IVM algorithm\footnote{\url{http://www.ipb.uni-bonn.de/ivm/}} is implemented in MATLAB and C++. 
The SVM classification is performed in MATLAB, using the LIBSVM approach by Chang and Lin \cite{Chang2001}.
To compute the result in (\ref{eq:DRF}) we use the graph-cut algorithm\footnote{\url{http://vision.csd.uwo.ca/code/}}~\cite{Boykov2001}.
Besides the standard SVM classification, we use the method of \cite{Platt2000} to convert the output of the SVM to probabilities (from now on referred to as probabilistic SVM). 

\section{Experimental Results}
\label{sec:discussion}
\begin{table*}[t]
\tabcolsep2pt
\centering
\caption{Overall accuracy (OA), average accuracy (AA) and kappa coefficient (Kappa) of SVM, SVM with transformed probabilities (Prob. SVM), IVM, incremental IVM (iIVM) with self-training (ST) and additional DRF. For the smaller data sets (10\%) we report the mean and standard deviation in brackets over 20 runs.}
\begin{tabular}{l|l|c|c|c|c|c|c|c|c|c}
	& & \multicolumn{3}{c|}{\city} & \multicolumn{3}{c|}{\uni}  & \multicolumn{3}{c}{\indian}\\
	Train data & Algorithm      & OA[\%] & AA[\%] & Kappa  &  OA[\%] & AA[\%] & Kappa & OA[\%] & AA[\%] & Kappa \\
	\hline\hline
	100\% & SVM     		& 97.1 & 94.6 & 0.95 & 79.0 & 87.9 & 0.73 & 61.6 & 69.6 & 0.57\\
	100\% & Prob. SVM     	& 97.0 & 94.2 & 0.95 & 78.4 & 87.7 & 0.73 & 60.7 & 68.9 & 0.56\\
	100\% & IVM     		& 97.2 & 94.2 & 0.95 & 78.3 & 87.4 & 0.73 & 63.7 & 67.8 & 0.59\\
	100\% & iIVM+ST 			& 97.3 & 95.2 & 0.95 & 86.6 & 90.0 & 0.83 & 63.7 & 67.8 & 0.59\\
	\hline
	100\% & Prob. SVM+DRF 	& 97.8 & 96.4 & 0.96 & 82.6 & 90.5 & 0.78 & 66.3 & 71.6 & 0.62 \\
	100\% & IVM+DRF 		& 98.4 & 96.5 & 0.97 & 85.9 & 91.4 & 0.82 & 70.6 & 80.0 & 0.67 \\
	100\% & iIVM+ST+DRF 	    & 98.6 & 97.1 & 0.98 & 94.2 & 93.8 & 0.92 & 70.6 & 80.2 & 0.67\\
	\hline
	10\% & SVM     			& 92.1 (0.6) & 85.1 (1.0) & 0.86 (0.01) & 70.2 (3.8) & 77.5 (2.4) & 0.62 (0.04) & 54.7 (2.3) & 63.0 (1.6) & 0.49 (0.03)\\
	10\% & Prob. SVM     	& 93.0 (1.3) & 86.2 (2.1) & 0.87 (0.02) & 69.4 (3.8) & 77.3 (2.4) & 0.61 (0.04) & 54.9 (2.5) & 63.2 (2.3) & 0.50 (0.03)\\
	10\% & IVM     			& 92.2 (1.1) & 85.4 (1.6) & 0.86 (0.02) & 70.3 (3.5) & 77.3 (0.8) & 0.62 (0.04) & 58.2 (0.1) & 65.2 (0.8) & 0.53 ($<$0.01)\\
	10\% & iIVM+ST 			& 94.3 (1.0) & 88.2 (1.7) & 0.90 (0.02) & 74.3 (2.4) & 77.9 (1.1) & 0.66 (0.02) & 64.8 (1.1) & 69.2 (1.1) & 0.61 (0.01)\\
	\hline
	10\% & Prob. SVM+DRF 	& 93.7 (1.5) & 88.9 (2.5) & 0.89 (0.03) & 76.3 (4.9) & 79.2 (4.5) & 0.70 (0.06) & 60.1 (2.8) & 65.9 (1.3) & 0.55 (0.03)\\
	10\% & IVM+ DRF 		& 92.8 (1.3) & 87.8 (2.0) & 0.87 (0.02) & 78.5 (4.3) & 79.9 (1.5) & 0.72 (0.05) & 64.5 (1.3) & 73.5 (2.0) & 0.60 (0.02)\\
	10\% & iIVM+ST+DRF 		& 95.8 (0.8) & 92.3 (2.1) & 0.92 (2.2) & 80.9 (2.9) & 81.0 (1.9) & 0.74 (0.03) & 70.1 (1.2) & 75.8 (2.5) & 0.66 (0.02)\\	
	
\end{tabular}
\label{tab:results}
\end{table*}

Fig. \ref{fig:classification} shows the ground truth and the classification results for all three data sets. 
As shown in Table \ref{tab:results}, the IVM is competitive to SVM in terms of accuracy and result in almost similar overall accuracies and kappa coefficients, irrespectively from the number of training samples. 
However, the SVM outperforms the IVM in terms of the averaged class accuracy (AA) for the \indian data set, when using all training samples. 
As expected the accuracies are increased by increasing the number of training samples, independently from the classifier method.

It is interesting to underline that in many cases, the probabilistic SVM provide (slightly) lower accuracies (i.e., OA and AA) than a standard SVM.
Behind this fact, the reliability of the probabilistic output of the SVM can be questioned. 
This assumption is confirmed by the results provided by the DRF. 
The accuracies of both methods, the probabilistic SVM and the IVM, are increased by the DRF.
This is in accordance with other studies that  have successfully integrated spatial information when classifying hyperspectral imagery~\cite{Borges2011,Tarabalka2010,Li2010a,Zhong2010,Zhong2008a}. 
However, the improvement of the IVM is usually higher, sometimes to a degree that the combination of IVM and DRF outperforms the accuracy provided by the probabilistic SVM in combination with a DRF.

\begin{table*}[htbp]
\caption{Class-specific accuracies of the \uni data set of SVM, SVM with transformed probabilities (Prob. SVM), IVM, incremental IVM (iIVM) with self-training (ST) and additional DRF. The results are given in percent ([\%]).}
\begin{center}
\begin{tabular}{l|c|c|c|c|c|c|c}
Class & SVM  & Prob.SVM  & IVM  & iIVM+ST & Prob.SVM + DRF  & IVM+DRF & iIVM+ST+DRF  \\
\hline
\hline
\colorbox{asphalt}{\parbox[h][0.2em][c]{1em}{}} Asphalt & 85.4 & 83.3 & 83.8 & 84.3 & 94.7 & 95.5 & 94.1 \\ 
\colorbox{bare}{\parbox[h][0.2em][c]{1em}{}} Bare Soil & 93.7 & 93.6 & 89.1 & 93.9 & 97.7 & 98.5 & 98.2 \\ 
\colorbox{bitumen}{\parbox[h][0.2em][c]{1em}{}} Bitumen & 90.5 & 90.5 & 90.3 & 90.7 & 94.1 & 95.7 & 94.5 \\ 
\colorbox{gravel}{\parbox[h][0.2em][c]{1em}{}} Gravel & 68.8 & 71.2 & 69.8 & 68.3 & 65.5 & 60.7 & 67.2 \\ 
\colorbox{meadows}{\parbox[h][0.2em][c]{1em}{}} Meadows & 65.9 & 65.2 & 66.2 & 82.8 & 68.6 & 66.6 & 93.7 \\ 
\colorbox{metal}{\parbox[h][0.2em][c]{1em}{}} Metal Sheets & 99.4 & 99.5 & 99.6 & 99.4 & 99.8 & 99.9 & 99.7 \\ 
\colorbox{bricks}{\parbox[h][0.2em][c]{1em}{}} Bricks & 92.5 & 91.8 & 92.7 & 93.7 & 98.7 & 99.3 & 98.8 \\ 
\colorbox{shadow}{\parbox[h][0.2em][c]{1em}{}} Shadow & 97.5 & 97.0 & 98.7 & 99.5 & 97.7 & 99.6 & 100.0 \\ 
\colorbox{trees}{\parbox[h][0.2em][c]{1em}{}} Trees & 97.0 & 97.4 & 95.9 & 97.0 & 97.9 & 98.3 & 98.0
\end{tabular}
\end{center}
\label{fig:class}
\end{table*}

Table \ref{fig:class} shows the class-specific accuracies achieved in the \uni data set.
The results confirm the previous findings, \eg, SVM and IVM show similar overall accuracies.
While some classes are more accurately classified by SVM, IVM are more adequate for the separation of other classes.

\begin{figure*}[tb]
  \centering
  \subfigure[\city]{\includegraphics[width=0.32\textwidth]{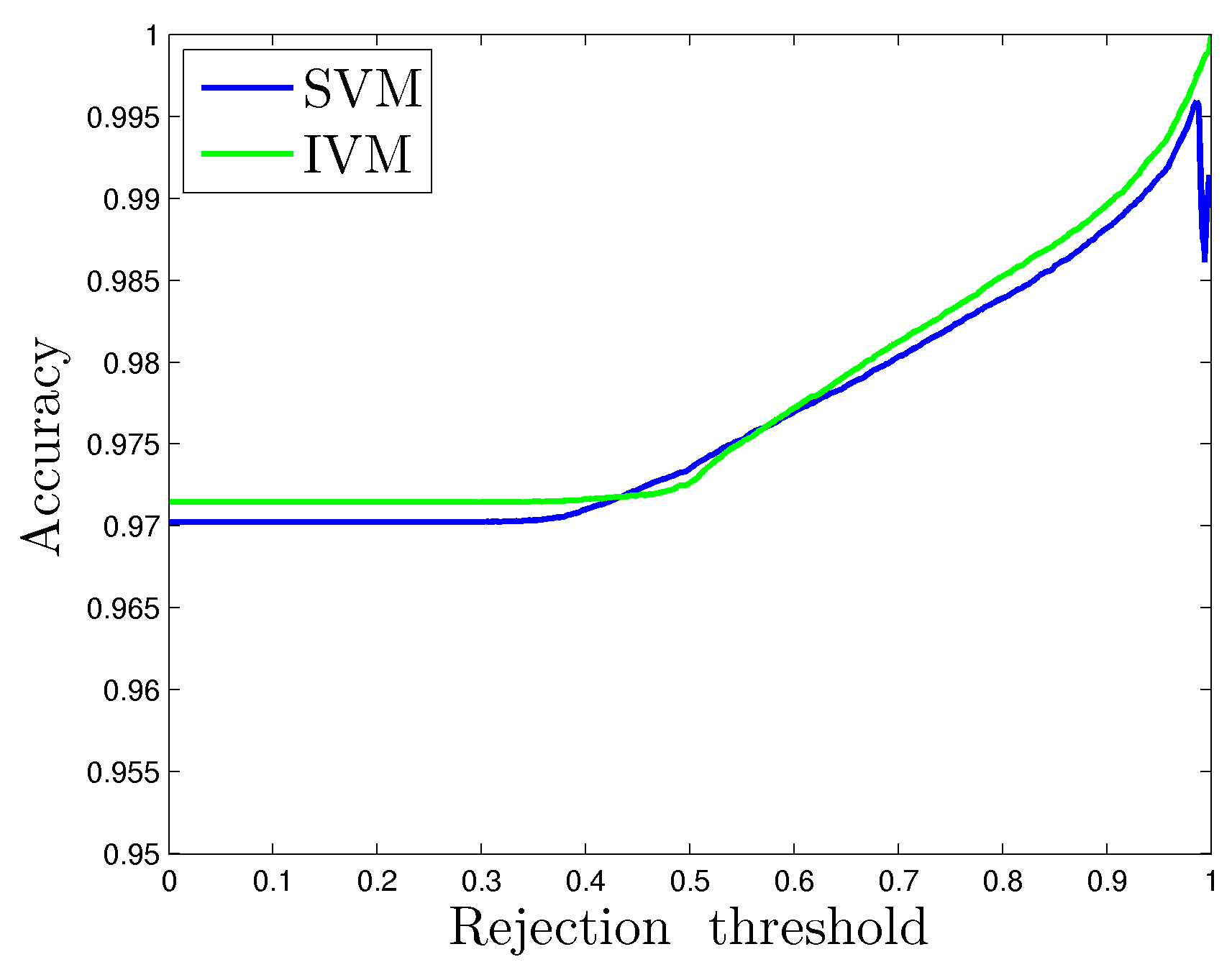}}
  \subfigure[\uni]{\includegraphics[width=0.32\textwidth]{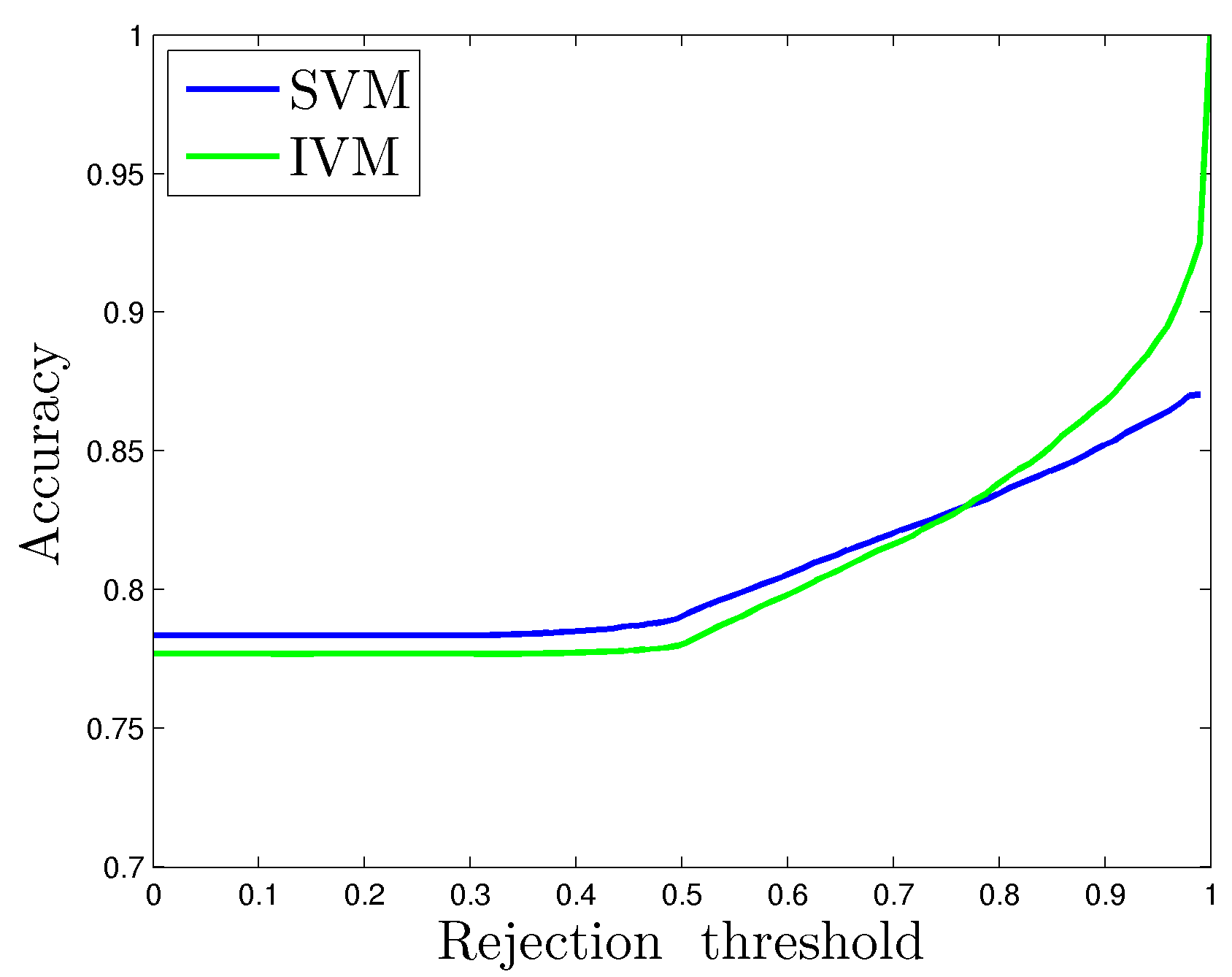}}
  \subfigure[\indian]{\includegraphics[width=0.32\textwidth]{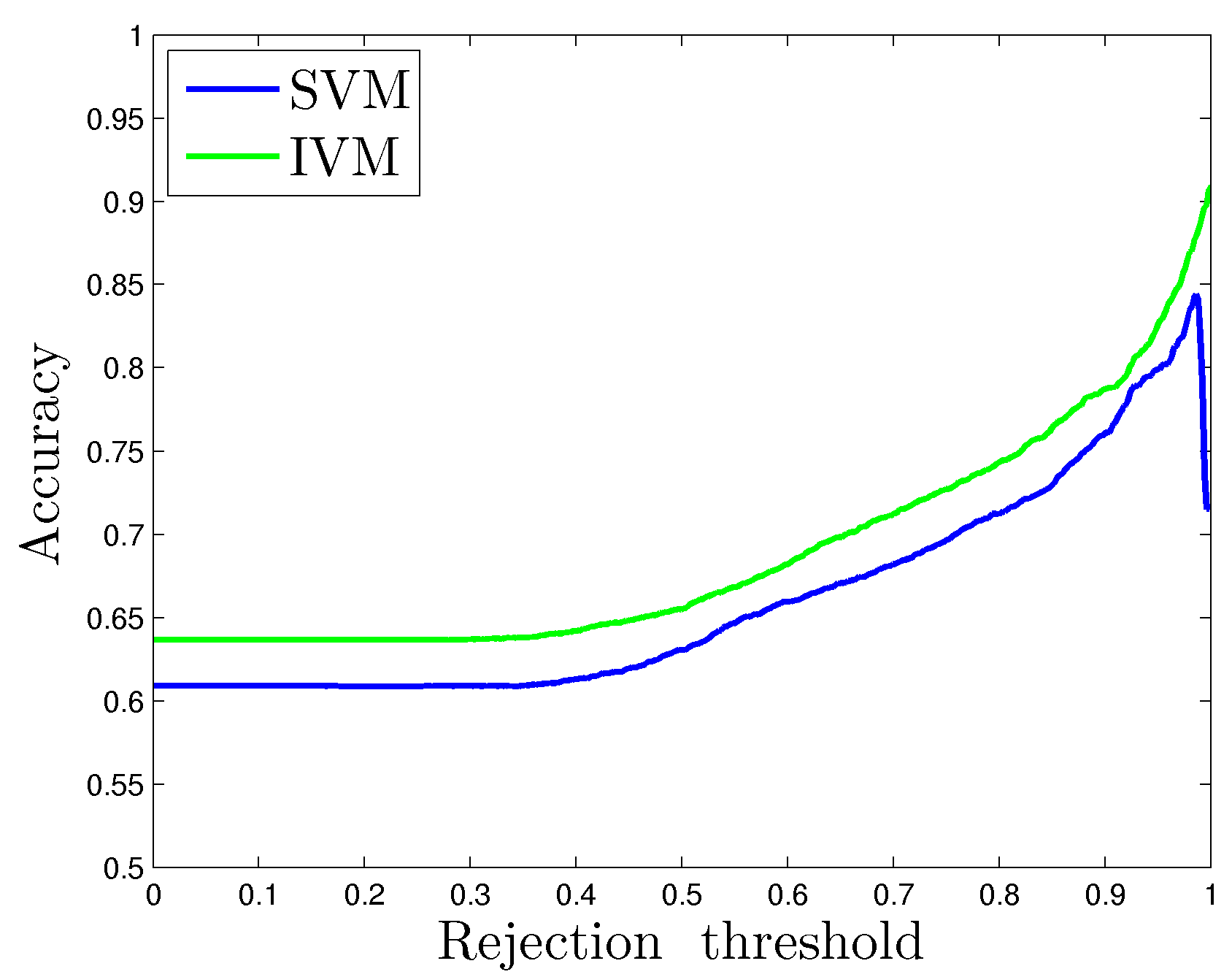}}
	\caption{The overall accuracy of SVM and IVM as a function of rejected test points on the \city data set (left), the \uni data set (middle) and the \indian data set (right). The overall accuracy is computed on the non-rejected test points.}
	\label{fig:rejection}
\end{figure*}

To underline this finding, an analysis of the probabilistic output is shown in Fig. \ref{fig:rejection}.
With an increasing rejection threshold, IVM provide higher OA on the \indian data and in most cases on the \city and \uni data set.
Consequently, it can be assumed that samples with high class probabilities are more accurately classified by IVM than by SVM, whereas relatively low class probabilities by the IVM are more likely referred to misclassified samples.

\begin{table}[tb]
\tabcolsep2pt
\centering
\scriptsize
\caption{Number of support/import vectors of SVM and IVM. For the smaller data sets (10\%) we report the mean and standard deviation in brackets over 20 runs.}
\begin{tabular}{l|l|c|c|c}
	Data & Algorithm & \city & \uni  & \indian\\
	\hline\hline
	100\% & SVM     		& 648 SV & 1111 SV & 2076 SV\\
	100\% & IVM     		& 30 IV & 109 IV  & 59 IV \\
	\hline
	10\% & SVM     			& 158.3 (11.8) SV &  209.9 (22.7) SV 	& 273.8 (10.1) SV\\
	10\% & IVM     			& 27.8 (4.9) IV  &   35.9 (6.0) IV  	& 75.0 (7.4) IV
\end{tabular}
\label{tab:resultsSparse}
\end{table}

Table \ref{tab:resultsSparse} shows, that the number of import vectors is lower when compared to the number of support vectors. 
This is in accordance with the results of a previous study in context of machine learning data sets \cite{Zhu2005}. 
Comparing the number of  support vectors and import vectors, respectively, the results confirm that the number of support vectors clearly increases with an increasing number of training samples, whereas the number of import vector increases slowly or almost remains constant.
Only for the \uni data set the number increases when all training points are used, because a smaller kernel parameter was chosen and more import vectors are necessary to train the classifier. 
Consequently, the computation time of the IVM during the classification is much faster when compared to SVM (see Table \ref{tab:resultsTime}), since the number of required mathematical operations depends on the number of support and import vectors.
This is particularly important in context of high-dimensional hyperspectral data sets, which are usually classified with a large number of training data.

\begin{table*}[tb]
\tabcolsep2pt
\centering
\caption{Training and test time of SVM, probabilistic SVM and IVM. For the smaller data sets (10\%) we report the mean and standard deviation in brackets over 20 runs.}
\begin{tabular}{l|l|c|c|c|c|c|c}
	& & \multicolumn{2}{c|}{\city} & \multicolumn{2}{c|}{\uni}  & \multicolumn{2}{c}{\indian}\\
	Data & Algorithm & Train[sec] & Test[sec] & Train[sec] & Test[sec]  & Train[sec] & Test[sec]  \\
	\hline\hline
	100\% & SVM     		& 3.8 & 187.9 & 4.3 & 126.4 & 20.0 & 68.9\\
	100\% & Prob. SVM     	& 106.1 &  0.1 & 41.7 & 0.1 & 12.45 &  0.1\\
	100\% & IVM     		& 225.2 & 2.3 & 484.9 & 1.8 & 344.3 & 0.1 \\
	\hline
	10\% & SVM     			& 0.1 ($<$ 0.1) & 48.2 (3.4) & 0.1 ($<$ 0.1) & 24.4 (2.5) & 0.3 ($<$ 0.1) & 6.2 (0.2)\\
	10\% & Prob. SVM     	& 105.0 (0.5) & 0.1 ($<$ 0.1) & 40.5 (0.25) & 0.1 ($<$ 0.1) & 12.4 (0.1) & 0.1 ($<$ 0.1)\\
	10\% & IVM     			& 11.4 (6.6) & 2.2 (0.2)  & 5.3 (1.9) & 0.9 (0.1) & 47.8 (10.0) & 0.1 ($<$ 0.1)\\
	
\end{tabular}
\label{tab:resultsTime}
\end{table*}

Table \ref{tab:resultsTime} reports the training and testing time of SVM and IVM on an Intel(R) Dual
Core with 3.0~GHz.
In contrast to the SVM implementation, the current Matlab/C++ implementation of the IVM is not optimized, so there is still potential for acceleration.

\begin{table*}[tb]
\tabcolsep2pt
\centering
\caption{Number of added training samples in the self-training (ST) procedure and number of training and import vectors before and after self-training. The number of training samples after self-training is given by the number of training samples before self-training plus the number of added training samples minus the removed irrelevant training samples. For the smaller data sets (10\%) we report the mean and standard deviation in brackets over 20 runs.}
\begin{tabular}{l|c|c|c|c|c|c}
	& \multicolumn{2}{c|}{\city} & \multicolumn{2}{c|}{\uni}  & \multicolumn{2}{c}{\indian}\\
	& 100\% & 10\% & 100\% & 10\% & 100\% & 10\% \\
	\hline\hline
	\# Training samples			& 5536 	& 556 & 3921 & 392 & 4811 & 563 \\
	\# Import vectors     		& 30 	& 27.8 (4.9) & 109 & 35.9 (6.0) & 59 & 75.0 (7.4) \\
	\# Added training samples	& 2610 	& 563.0 (158.7) & 2340 & 1390.0 (463.5) & 16 & 584.0 (267.4)\\
	\# Training samples (ST) & 538 	& 493.8 (108.2) & 4282 & 1208.6 (439.1) & 4719 & 911.4 (139.9)\\
	\# Import vectors (ST)	& 21 	& 29.7  (2.6) & 49 & 42.7 (5.7) & 61 & 67.1 (14.7)
\end{tabular}
\label{tab:resultsSelf}
\end{table*}

Finally, Table \ref{tab:results} also shows the positive impact of self-training on the classification accuracy.
The classification result was improved in all cases. 
As expected the improvement on the training sets with 10\% of all training samples is higher when compared to the classification results, which were generated by the whole training samples set. 
Table \ref{tab:resultsSelf} shows, that the number of import vectors remains nearly the same during the self-training procedure.
Moreover, the proposed self-training strategy deletes irrelevant training samples. 
Therefore, the final number of training samples is significantly lower, when compared to initial number of training samples and the samples added during the self-training procedure. This fact is particularly obvious in case of the \city data using the whole training sample set.

\begin{figure*}[tb]
  \centering
  \subfigure[Data]{
  \begin{minipage}{0.125\textwidth}
  \frame{\includegraphics[width=1\textwidth]{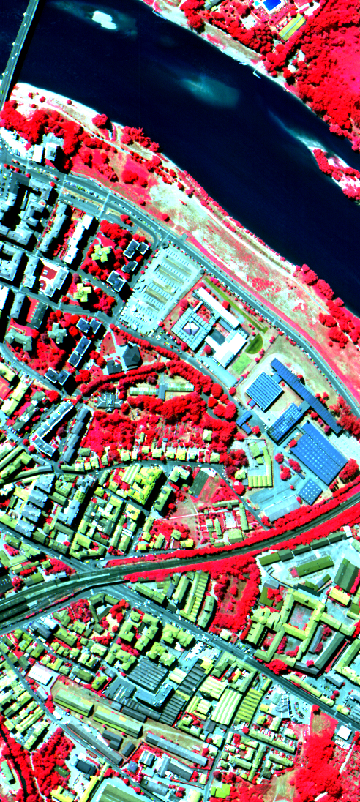}}\\[0.5em]
  \frame{\includegraphics[width=1\textwidth]{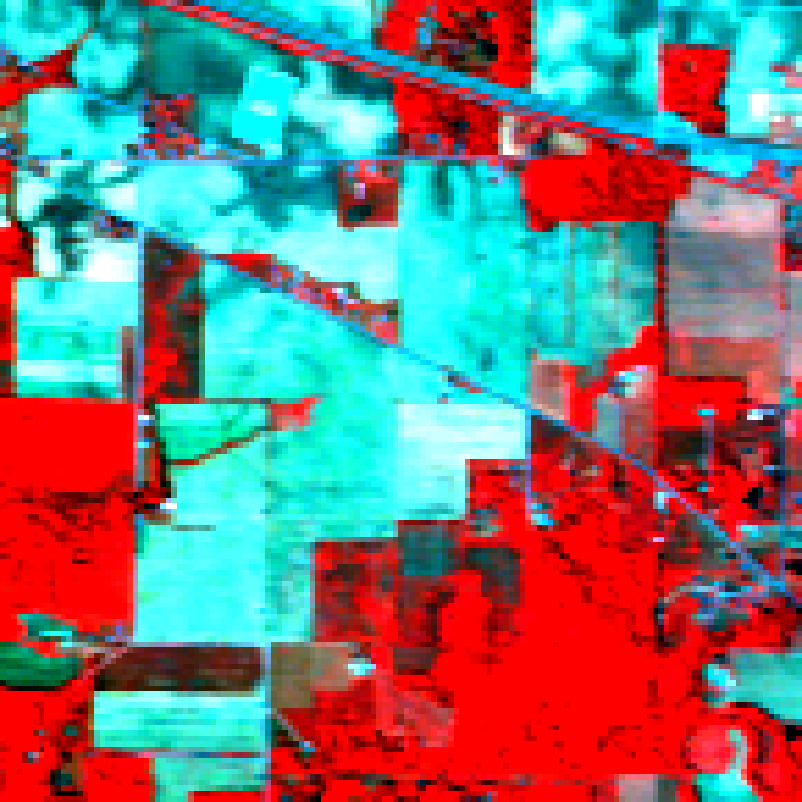}}\\[0.5em] 
  \frame{\includegraphics[width=1\textwidth]{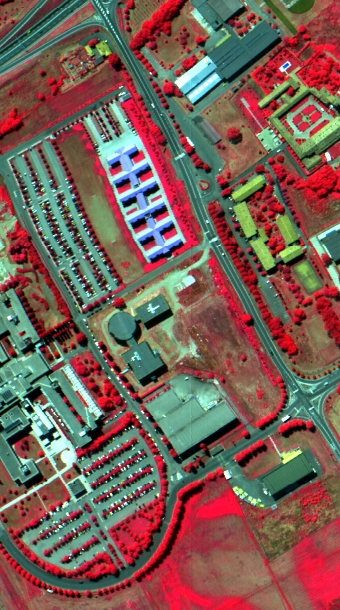}}\\
  \end{minipage}}
  \subfigure[Training ground truth]{
  \begin{minipage}{0.125\textwidth}
  \frame{\includegraphics[width=1\textwidth]{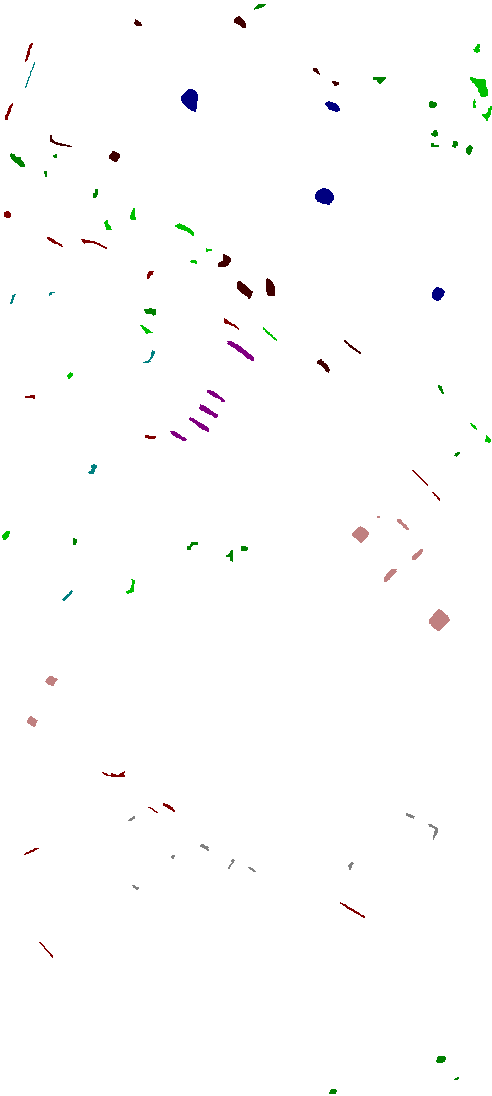}}\\[0.5em]
  \frame{\includegraphics[width=1\textwidth]{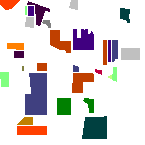}}\\[0.5em]
  \frame{\includegraphics[width=1\textwidth]{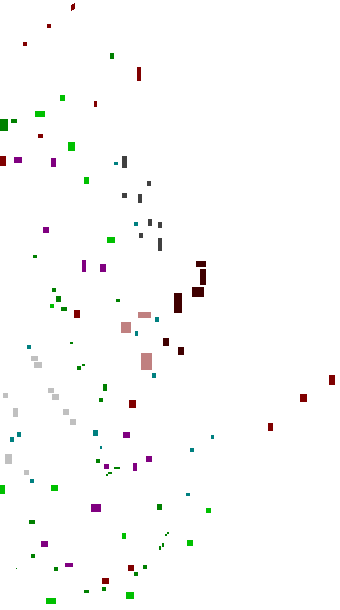}}\\
  \end{minipage}}
  \subfigure[Test ground truth]{
  \begin{minipage}{0.125\textwidth}
  \frame{\includegraphics[width=1\textwidth]{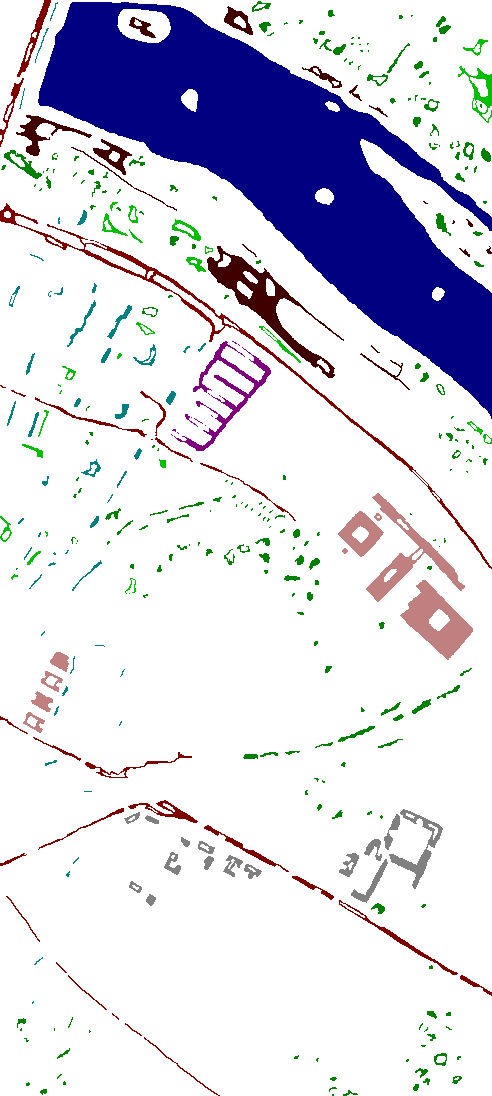}}\\[0.5em]
  \frame{\includegraphics[width=1\textwidth]{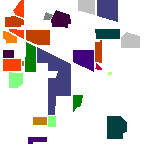}}\\[0.5em]
  \frame{\includegraphics[width=1\textwidth]{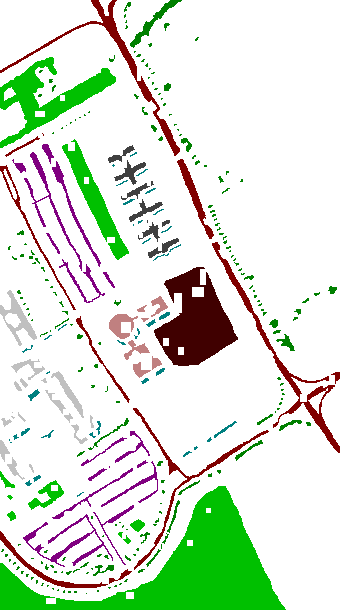}}\\
  \end{minipage}}
  \subfigure[SVM result]{
  \begin{minipage}{0.125\textwidth}
  \frame{\includegraphics[width=1\textwidth]{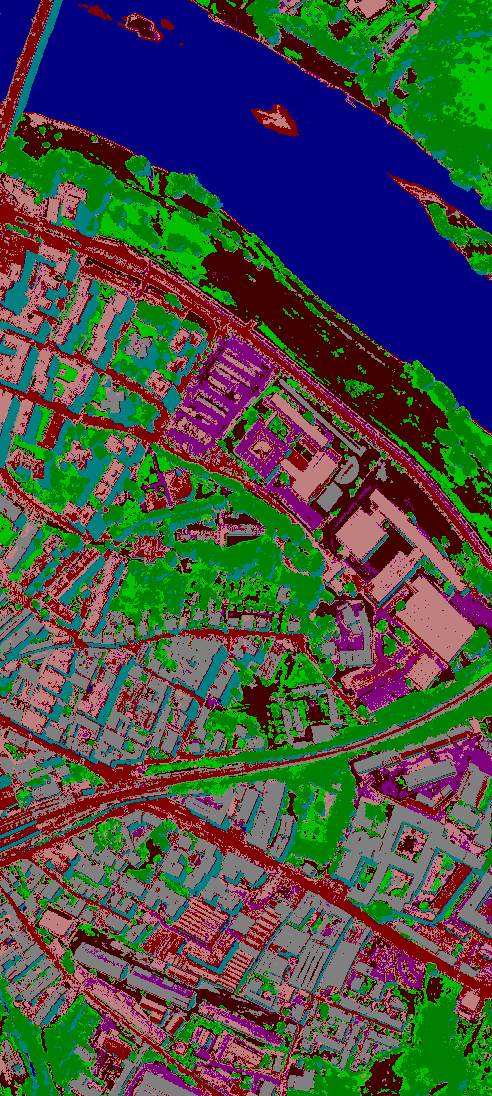}}\\[0.5em]
  \frame{\includegraphics[width=1\textwidth]{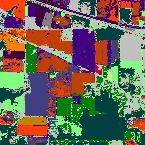}}\\[0.5em]
  \frame{\includegraphics[width=1\textwidth]{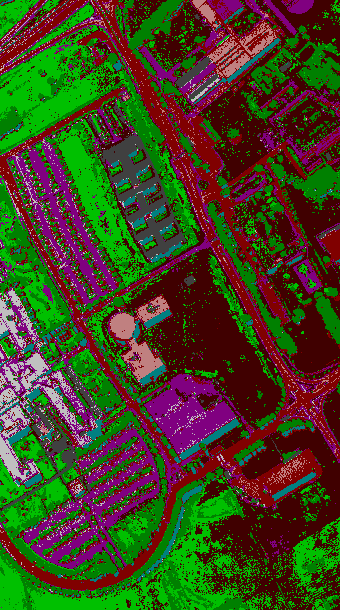}}\\
  \end{minipage}}
  \subfigure[IVM result]{
  \begin{minipage}{0.125\textwidth}
  \frame{\includegraphics[width=1\textwidth]{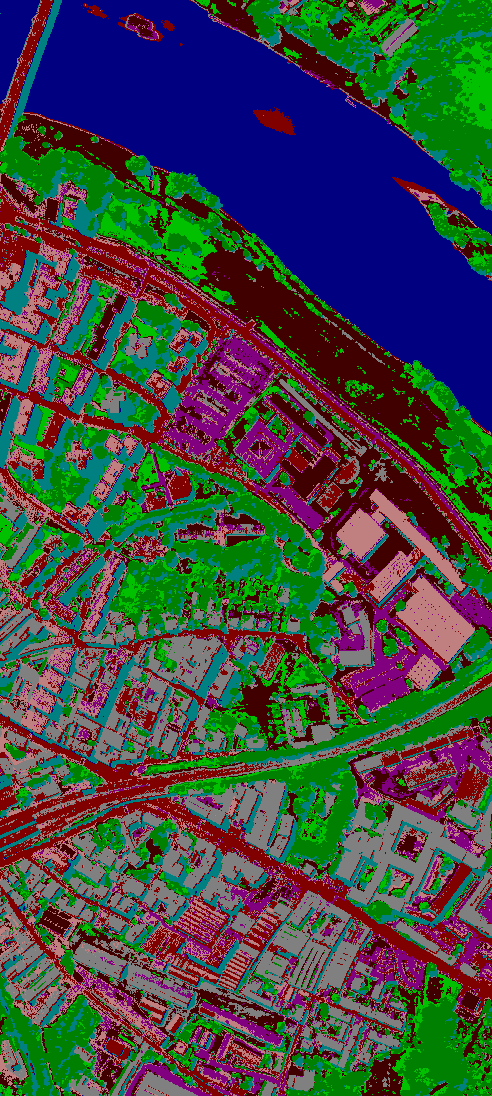}}\\[0.5em]
  \frame{\includegraphics[width=1\textwidth]{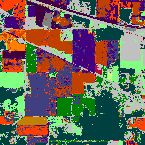}}\\[0.5em]
  \frame{\includegraphics[width=1\textwidth]{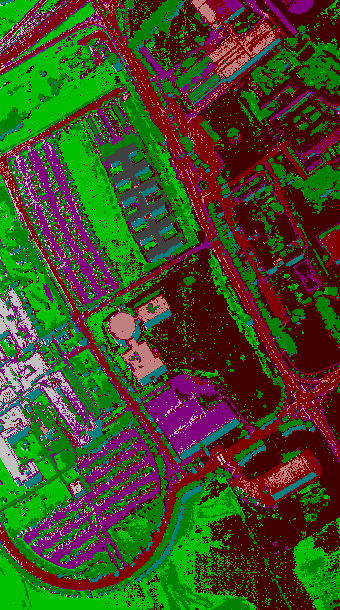}}\\
  \end{minipage}}
  \subfigure[IVM+DRF result]{
  \begin{minipage}{0.125\textwidth}
  \frame{\includegraphics[width=1\textwidth]{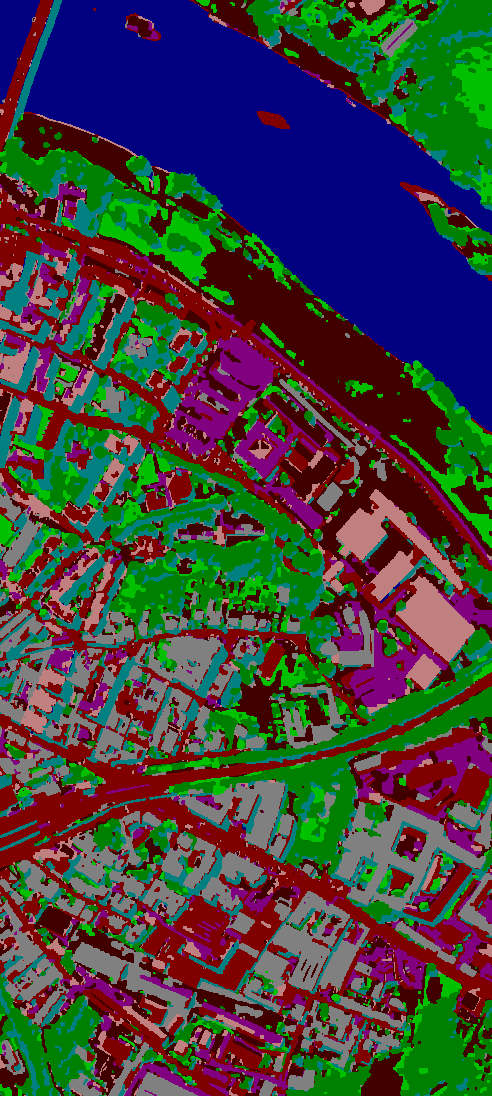}}\\[0.5em]
  \frame{\includegraphics[width=1\textwidth]{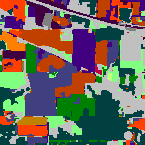}}\\[0.5em]
  \frame{\includegraphics[width=1\textwidth]{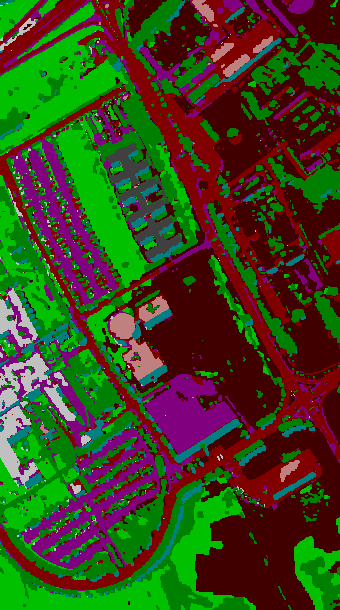}}\\
  \end{minipage}}
  \subfigure[Self-training result]{
  \begin{minipage}{0.125\textwidth}
  \frame{\includegraphics[width=1\textwidth]{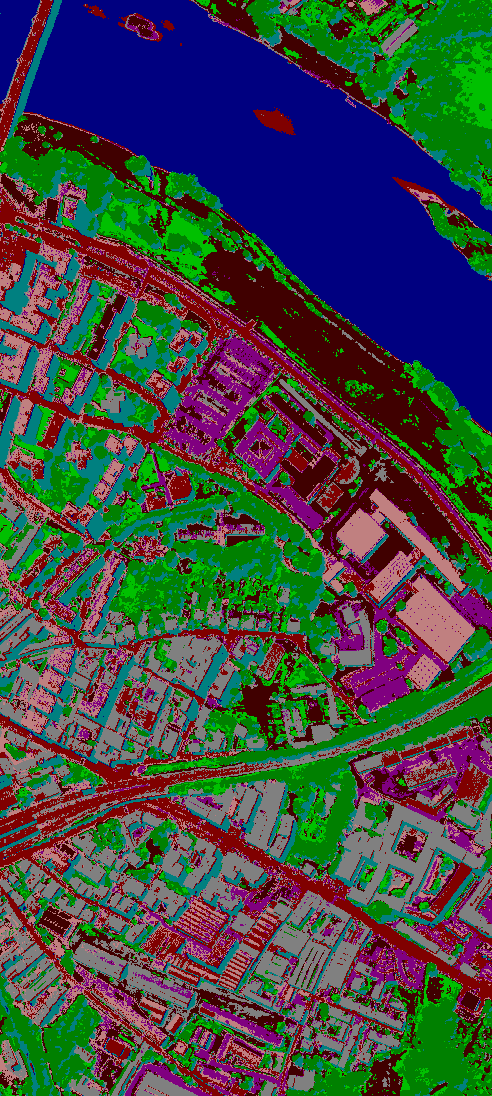}}\\[0.5em]
  \frame{\includegraphics[width=1\textwidth]{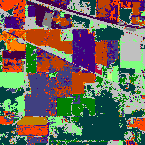}}\\[0.5em]
  \frame{\includegraphics[width=1\textwidth]{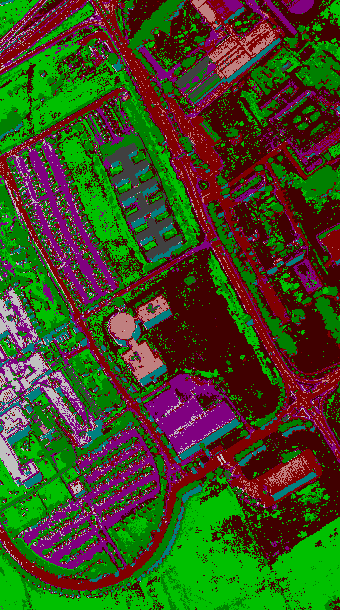}}\\
  \end{minipage}}
	\caption{F.l.t.r.: (a) Data, (b) training ground truth, (c) test ground truth, (d) classification result of SVM, (e) IVM, (f) IVM + DRF and (g) self-training procedure with incremental IVM. The upper row shows the \city data set, the middle row the \indian data set and the bottom row the \uni data set.}
	\label{fig:classification}
\end{figure*}

% -------------------------------------------------------%
\section{Conclusion and Outlook}
\label{sec:conclusion}
We proposed the incremental IVM classifier, which includes the addition and deletion of training samples as well as the update of the set of import vectors.
The incremental learning strategy updates efficiently the classifier model without re-training from scratch, which makes it capable for large data sets.
To evaluate the incremental IVM, we have introduced a self-training strategy, which uses the probabilistic output of the classifier and a DRF.

We evaluated the performance of IVM in the context of classifying hyperspectral imagery. 
IVM constitute a feasible approach and an useful alternative for the classification of remote sensing data, particularly when probabilities are of interest.
The experimental results underline that SVM and IVM perform almost similar in terms of the classification accuracy. 
In addition, the results show the strong dependency of the number of support vectors on the number of available training samples. In contrast to this, the number of import vectors is significantly lower when compared to the number of support vectors and remains constant or only slightly increases with an increasing number of training samples. 
As confirmed by the experimental results, the probabilities provided by IVM are more reliably, when compared to the probabilistic outputs provided by SVM. 
This fact is particularly interesting, because the probabilities are useful for further image analysis, e.g., (i) as input in a DRF that increases the classification accuracy, (ii) to detect mislabeled samples by a uncertainty analysis, and (iii) to identity relevant training samples for a self-training strategy.
Particularly for hyperspectral data sets, which require a sufficient large number of training samples to ensure an adequate accuracy, the self-training strategy including the incremental IVM is interesting  and can further increase the classification result. Moreover, the computation time is reduced by the incremental learning approach rather than re-training the classifier with all training samples. The incremental IVM can further be incorporated into other active learning approaches or more sophisticated models for DRFs. Therefore, the approach seems attractive as well as feasible for operational applications.

Overall, the IVM and its incremental version appears worthwhile for the classification of remote sensing data, especially when the user is interested in reliable class probabilities and a fast classification. 
More efficient implementation strategies and further modifications will be investigated in the future. 
 
%%%%%%%%%%%%%%%%%%%%%%%%%%%%%%%%%%%%%%%%%%%%%%%%%%%%%%%%%%%%%%%%%%%%%%%%%
\ifCLASSOPTIONcompsoc
  \section*{Acknowledgments}
\else
  \section*{Acknowledgment}
\fi
The authors would like to thank D.~Landgrebe and L.~Biehl~(Purdue University, USA) for providing the Indian Pines data~(available on: \url{http://cobweb.ecn.purdue.edu/~biehl/MultiSpec/}) and P.~Gamba~(University of Pavia, Italy) for providing the Pavia dataset.

\ifCLASSOPTIONcaptionsoff
  \newpage
\fi
\bibliographystyle{IEEEtran}

\newpage
\begin{IEEEbiography}[{\includegraphics[width=1in,height=1.25in,clip,keepaspectratio]{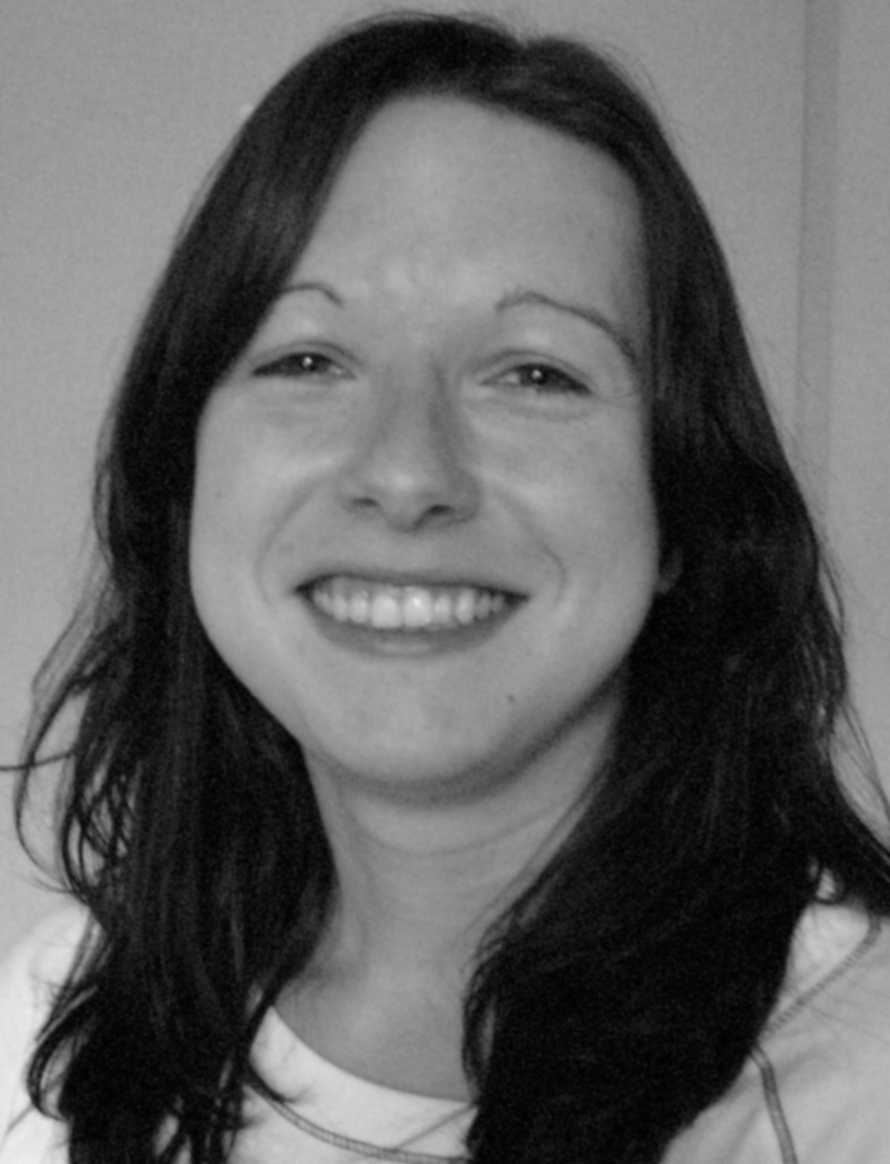}}]{Ribana Roscher}
received her Dipl.-Ing. degree in Geodesy from University of Bonn, Germany, in 2008.
She is currently a PhD at the Institute of Geodesy and Geoinformation at the University of Bonn, Germany.
Her current research activities concentrate on sequential learning and discriminative models for semantic segmentation, especially import vector machines.
She is reviewer for IEEE Transactions on Geoscience and Remote Sensing and IEEE Transactions on Pattern Analysis and Machine Intelligence.
\end{IEEEbiography}

\begin{IEEEbiography}[{\includegraphics[width=1in,height=1.25in,clip,keepaspectratio]{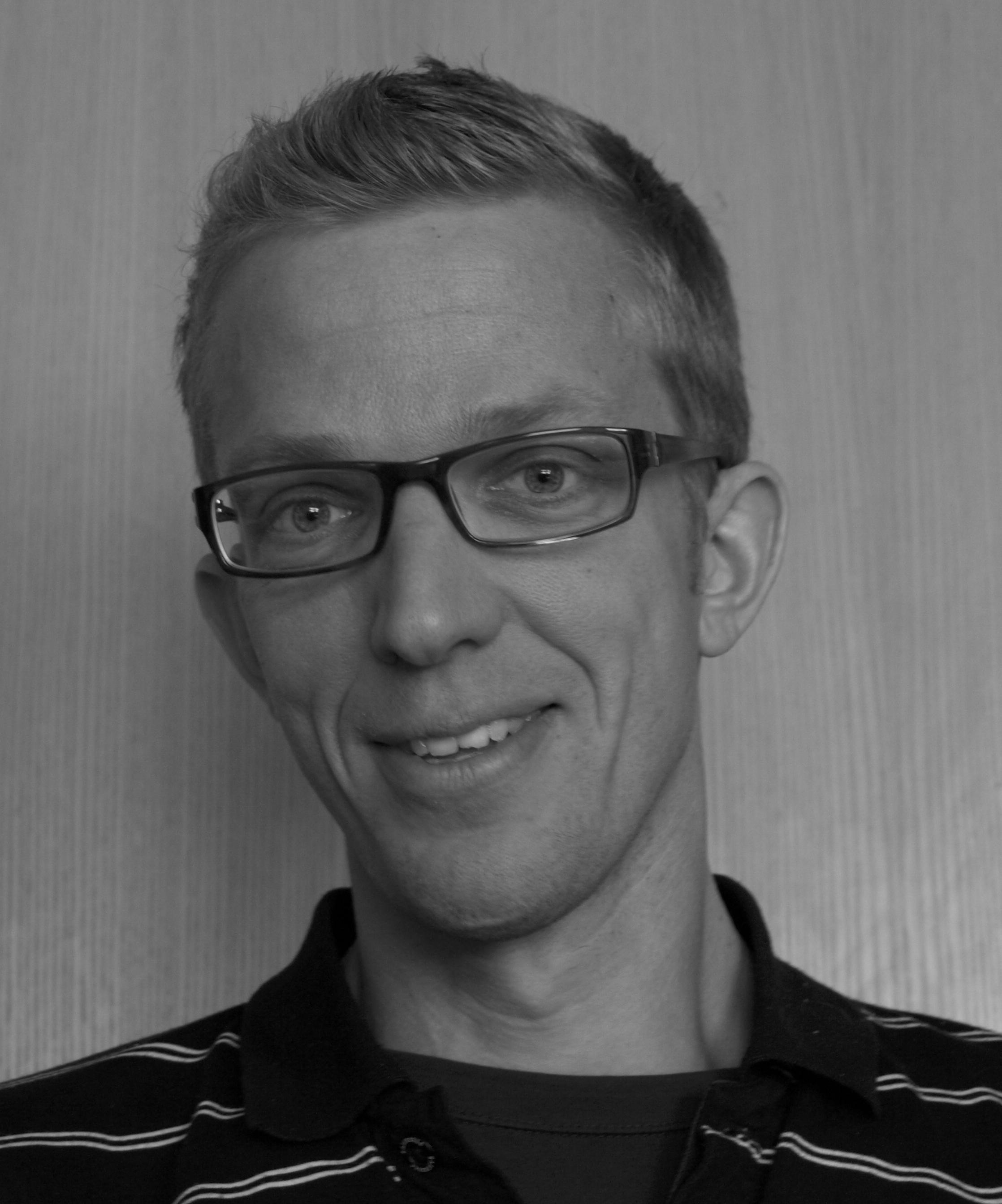}}]
{Bj\"orn Waske} (S06-M08) received his degree in Applied Environmental Sciences with a major in Remote Sensing from Trier University, Germany, in 2002. Until mid 2004 he was research assistant the Department of Geosciences at the Munich University, Germany. From 2004 until end of 2007 he pursued a PhD at the Center for Remote Sensing of Land Surfaces (ZFL) at the University of Bonn, Germany and received the PhD degree in Geography. From beginning of 2008 until August 2009 he was a Postdoctoral researcher at the Faculty of Electrical and Computer Engineering, University of Iceland. Since September 2009 he is a (Junior)professor for Remote Sensing in Agriculture at the University of Bonn, Germany. His current research activities concentrate on advanced concepts for image classification and data fusion.
Currently he is Associate Editor IEEE Journal of Selected Topics in Applied Earth Observations and Remote Sensing (J-STARS). He is reviewer for different international journals, including IEEE Transactions on Geoscience and Remote Sensing, IEEE Geoscience and Remote Sensing Letters, and IEEE Journal of Selected Topics in Applied Earth Observations and Remote Sensing.
\end{IEEEbiography}

\begin{IEEEbiography}[{\includegraphics[width=1in,height=1.25in,clip,keepaspectratio]{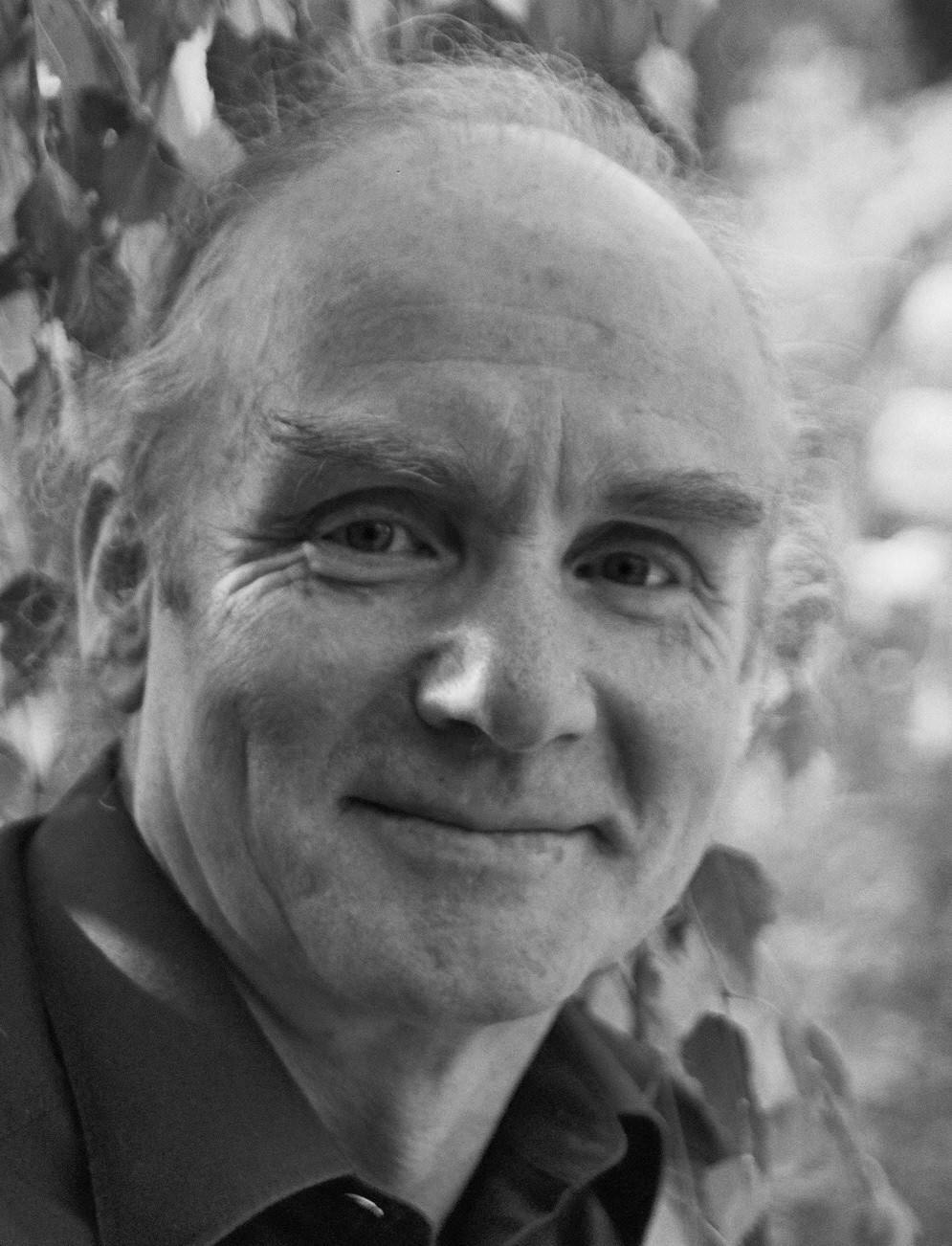}}]
{Wolfgang F\"orstner}, born 1946, received his Dipl.-Ing. degree in Geodesy at the University of Stuttgart, Germany, in 1971. He received the PhD degree in 1976 at the University of Stuttgart, Germany. Since 1990 he chairs the Department of Photogrammetry at the University of Bonn, Germany. His fields of interest are digital photogrammetry, statistical methods of image analysis, analysis of image sequences, semantic modeling, machine learning and geoinformation systems. He published more than 130 scientific papers, supervised appr. 70 Bachelor and Master Theses and more than 30 PhD Theses. From 1994-2001 he was vice president of the German Association for Pattern Recognition (DAGM). He currently is associated editor of IEEE Transactions on Pattern Analysis and Machine Intelligence. 
\end{IEEEbiography}

\vfill

\end{document}